\documentclass[final]{cvpr}

\usepackage{times}
\usepackage{epsfig}
\usepackage{graphicx}
\usepackage{amsmath}
\usepackage{amssymb}

\usepackage{enumitem}
\usepackage{booktabs}

\usepackage{pgfplots} %
\usepackage{pgfplotstable} %
\pgfplotsset{compat=newest} %
\usetikzlibrary{plotmarks} %
\usetikzlibrary{colorbrewer} %

\usepackage{xspace}

\usepackage{algorithm}
\usepackage{algpseudocode}
\usepackage{multirow}
\usepackage{tabularx}
\usepackage{wrapfig}
\usepackage{stfloats}  %

\makeatletter
\DeclareRobustCommand\onedot{\futurelet\@let@token\@onedot}
\def\@onedot{\ifx\@let@token.\else.\null\fi\xspace}

\usepackage[pagebackref=true,breaklinks=true,colorlinks,bookmarks=false]{hyperref}

\def\eg{\emph{e.g}\onedot} 
\def\ie{\emph{i.e}\onedot} 
 
 \def\vs{\emph{vs}\onedot}
 
\def\etal{\emph{et al}\onedot}

\usepackage{pifont}
\newcommand{\cmark}{\ding{51}}

\def\cvprPaperID{1280} %
\def\confYear{CVPR 2021}

\begin{document}

\title{Scaling Wide Residual Networks for Panoptic Segmentation}

\author{Liang-Chieh Chen\textsuperscript{1}~~~~~~Huiyu Wang\textsuperscript{2}\thanks{Work done while an intern at Google.}~~~~~~Siyuan Qiao\textsuperscript{2}\footnotemark[1]\\
\textsuperscript{1}Google Research\\\textsuperscript{2}Johns Hopkins University\\
}

\maketitle

\begin{abstract}
The Wide Residual Networks (Wide-ResNets), a shallow but wide model variant of the Residual Networks (ResNets) by stacking a small number of residual blocks with large channel sizes, have demonstrated outstanding performance on multiple dense prediction tasks. However, since proposed, the Wide-ResNet architecture has barely evolved over the years. In this work, we revisit its architecture design for the recent challenging panoptic segmentation task, which aims to unify semantic segmentation and instance segmentation. A baseline model is obtained by incorporating the simple and effective Squeeze-and-Excitation and Switchable Atrous Convolution to the Wide-ResNets. Its network capacity is further scaled up or down by adjusting the width (\ie, channel size) and depth (\ie, number of layers), resulting in a family of SWideRNets (short for Scaling Wide Residual Networks). We demonstrate that such a simple scaling scheme, coupled with grid search, identifies several SWideRNets that significantly advance state-of-the-art performance on panoptic segmentation datasets in both the fast model regime and strong model regime.
\end{abstract}

\section{Introduction}
\label{sec:intro}

\begin{figure}
    \hspace{-4mm}
    \begin{tabular} {c}
    \vspace{-2mm}
    \resizebox{\linewidth}{!}{\begin{tikzpicture}[/pgfplots/width=1\linewidth, /pgfplots/height=0.64\linewidth, /pgfplots/legend pos=south east]
    \begin{axis}[
        ymin=25,ymax=45,xmin=20,xmax=180,
        xlabel=inference time (ms),
        ylabel=PQ (\%),
		font=\scriptsize,
        grid=both,
		grid style=dotted,
        legend columns=1,
        xlabel shift={-2pt},
        ylabel shift={-5pt},
	    xtick={0,50,...,300},
	    ytick={0,5,...,60},
        legend pos= outer north east,
        legend cell align={left},
        legend style={font=\scriptsize},
        ]

	    \addplot[orange,mark=*,only marks,mark size=1.75] coordinates{(30, 31.5)};
        \addlegendentry{\hphantom{i}SWideRNet-(0.25, 0.25, 0.75)}
        
        \addplot[green,mark=*,only marks,mark size=1.75] coordinates{(31.45, 34.3)};
        \addlegendentry{\hphantom{i}SWideRNet-(0.25, 0.35, 0.75)}
        
        \addplot[blue,mark=*,only marks,mark size=1.75] coordinates{(35.96, 36)};
        \addlegendentry{\hphantom{i}SWideRNet-(0.25, 0.35, 1)}
        
        \addplot[cyan,mark=*,only marks,mark size=1.75] coordinates{(42.57, 38.1)};
        \addlegendentry{\hphantom{i}SWideRNet-(0.25, 0.5, 1)}
        
        \addplot[red,mark=*,only marks,mark size=1.75] coordinates{(56.66, 40.1)};
        \addlegendentry{\hphantom{i}SWideRNet-(0.25, 0.75, 1)}
        
	    \addplot[brown,mark=x,only marks,mark size=1.75] coordinates{(74, 38.9)};
        \addlegendentry{\hphantom{i}Panoptic-DeepLab (X-71) \cite{cheng2019panoptic}}
    
        \addplot[brown,mark=triangle,only marks,mark size=1.75] coordinates{(50, 35.1)};
        \addlegendentry{\hphantom{i}Panoptic-DeepLab (R-50) \cite{cheng2019panoptic}}

        \addplot[brown,mark=o,only marks,mark size=1.75] coordinates{(38, 30.0)};
        \addlegendentry{\hphantom{i}Panoptic-DeepLab (MNV3) \cite{cheng2019panoptic}}

        \addplot[teal,mark=asterisk,only marks,mark size=1.75] coordinates{(119, 33.8)};
        \addlegendentry{\hphantom{i}DeeperLab (X-71) \cite{yang2019deeperlab}}
       
        \addplot[teal,mark=square,only marks,line width=1, mark size=1.75] coordinates{(83, 27.9)};
        \addlegendentry{\hphantom{i}DeeperLab (WMNV2) \cite{yang2019deeperlab}}

        \addplot[olive,mark=star, mark size=1.75,only marks, line width=1] coordinates{(167, 42.5)};
        \addlegendentry{\hphantom{i}UPSNet (R-50) \cite{xiong2019upsnet}}
        
        \addplot[gray,mark=diamond, mark size=1.75,only marks, line width=1] coordinates{(176.5, 37.5)};
        \addlegendentry{\hphantom{i}PCV (R-50) \cite{Wang_2020_CVPR}}
        
        \addplot[lightgray,mark=pentagon, mark size=1.75,only marks, line width=1] coordinates{(63, 37.1)};
        \addlegendentry{\hphantom{i}Hou (R-50) \etal \cite{hou2020real}}
        
    \end{axis}
\end{tikzpicture}} \\
    {\scriptsize (a) PQ \vs inference time (ms) on COCO val set.} \\
    \vspace{-2mm}
    \resizebox{\linewidth}{!}{\begin{tikzpicture}[/pgfplots/width=1\linewidth, /pgfplots/height=0.64\linewidth, /pgfplots/legend pos=south east]
    \begin{axis}[
        ymin=50,ymax=65,xmin=0,xmax=500,
        xlabel=inference time (ms),
        ylabel=PQ (\%),
		font=\scriptsize,
        grid=both,
		grid style=dotted,
        legend columns=1,
	    xtick={0,50,...,500},
	    ytick={0,5,...,70},
        legend pos= outer north east,
        legend cell align={left}
        ]

        \addplot[orange,mark=*,only marks,mark size=1.75] coordinates{(63.05, 58.4)};
        \addlegendentry{\hphantom{i}SWideRNet-(0.25, 0.25, 0.75)}
        
        \addplot[green,mark=*,only marks,mark size=1.75] coordinates{(73.14, 59.8)};
        \addlegendentry{\hphantom{i}SWideRNet-(0.25, 0.35, 0.75)}
        
        \addplot[blue,mark=*,only marks,mark size=1.75] coordinates{(83.25, 61.6)};
        \addlegendentry{\hphantom{i}SWideRNet-(0.25, 0.35, 1)}
        
        \addplot[cyan,mark=*,only marks,mark size=1.75] coordinates{(108.58, 62.7)};
        \addlegendentry{\hphantom{i}SWideRNet-(0.25, 0.5, 1)}
        
        \addplot[red,mark=*,only marks,mark size=1.75] coordinates{(166.83, 63.8)};
        \addlegendentry{\hphantom{i}SWideRNet-(0.25, 0.75, 1)}
        
	    \addplot[brown,mark=x,only marks,mark size=1.75] coordinates{(175, 63)};
        \addlegendentry{\hphantom{i}Panoptic-DeepLab (X-71) \cite{cheng2019panoptic}}
        
        \addplot[brown,mark=triangle,only marks,mark size=1.75] coordinates{(117, 59.7)};
        \addlegendentry{\hphantom{i}Panoptic-DeepLab (R-50) \cite{cheng2019panoptic}}

        \addplot[brown,mark=o,only marks,mark size=1.75] coordinates{(63, 55.4)};
        \addlegendentry{\hphantom{i}Panoptic-DeepLab (MNV3) \cite{cheng2019panoptic}}

        \addplot[teal,mark=asterisk,only marks,mark size=1.75] coordinates{(463, 56.5)};
        \addlegendentry{\hphantom{i}DeeperLab (X-71) \cite{yang2019deeperlab}}
       
        \addplot[teal,mark=square,only marks,line width=1, mark size=1.75] coordinates{(303, 52.3)};
        \addlegendentry{\hphantom{i}DeeperLab (WMNV2) \cite{yang2019deeperlab}}

        \addplot[olive,mark=star, mark size=1.75,only marks, line width=1] coordinates{(202, 59.3)};
        \addlegendentry{\hphantom{i}UPSNet (R-50) \cite{xiong2019upsnet}}
        
        \addplot[gray,mark=diamond, mark size=1.75,only marks, line width=1] coordinates{(182.8, 54.2)};
        \addlegendentry{\hphantom{i}PCV (R-50) \cite{Wang_2020_CVPR}}
        
        \addplot[lightgray,mark=pentagon, mark size=1.75,only marks, line width=1] coordinates{(99, 58.8)};
        \addlegendentry{\hphantom{i}Hou (R-50) \etal \cite{hou2020real}}
    \end{axis}
\end{tikzpicture}} \\
    {\scriptsize (b) PQ \vs inference time (ms) on Cityscapes Vistas val set} \\
    \vspace{-2mm}
    \end{tabular}
    \caption{PQ \vs GPU inference time. Our fast SWideRNet model variants, deployed as the network backbone in  Panoptic-DeepLab~\cite{cheng2019panoptic}, significantly improve the speed-accuracy trade-off.
    }
    \vspace{-2mm}
    \label{fig:pq_vs_second}
\end{figure}
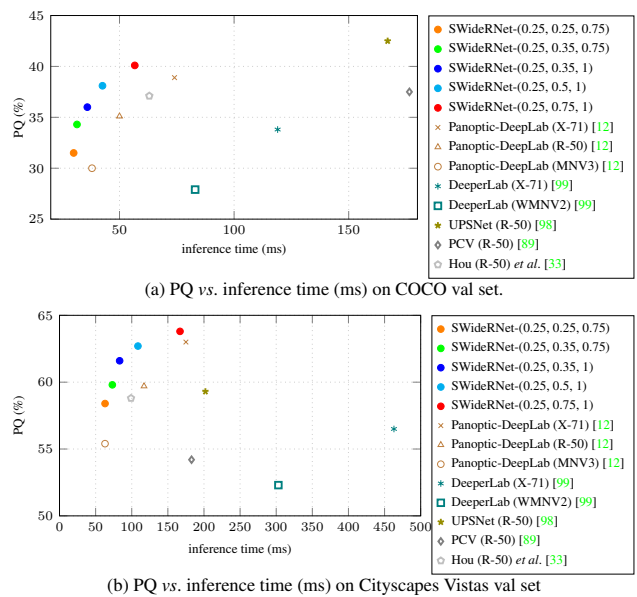

Computer vision systems have achieved remarkable performance across a wide range of image recognition tasks, including image classification~\cite{krizhevsky2012imagenet,simonyan2014very}, object detection~\cite{girshick2014rich,ren2015faster}, and dense prediction~\cite{long2015fully,deeplabv12015}, thanks to the recent advances in  learning algorithms~\cite{goodfellow2016deep} (\eg, better optimizer~\cite{kingma2015adam}, normalization techniques~\cite{ioffe2015batch,salimans2016weight,wu2018group,qiao2019weight}, and scalable training systems~\cite{tensorflow-osdi2016,goyal2017accurate,peng2018megdet}). The improvement of neural network architectures especially plays an important role, as manifested on public benchmarks~\cite{everingham2010pascal,lin2014microsoft,ILSVRC15}. 

Gaining in popularity for its simplicity and effectiveness, Residual Networks (ResNets)~\cite{he2016deep} have been the building blocks of many modern neural network architectures~\cite{wrn2016wide,xie2017aggregated,hu2018squeeze,wu2019wider,li2019selective,gao2019res2net,radosavovic2020designing,zhang2020resnest}. Specifically, the Wide-ResNets~\cite{wrn2016wide} adopt the `shallow but wide' architecture design (\ie, fewer layers but with large channels) and show superior performance over the `deep but thin' architectures (\ie, more layers but with small channels). Along the same direction, the Wide-ResNet-38 (WR-38)~\cite{wu2019wider}, a sophisticated human-designed wide residual network, is one of the top-performing network backbones on many dense prediction benchmarks~\cite{cordts2016cityscapes,neuhold2017mapillary}.
However, since proposed in 2016, the architecture of WR-38 has barely evolved over the years.
Recently, a simple modification of WR-38, by altering the last two residual blocks, leads to a slightly better and faster architecture WR-41~\cite{chen2020naive}, when deployed as the network backbone in  Panoptic-DeepLab~\cite{cheng2019panoptic} framework. The resulting model has shown state-of-the-art performance for panoptic  segmentation~\cite{kirillov2018panoptic}, which is a challenging dense prediction task with the goal to unify semantic segmentation~\cite{he2004multiscale} and instance segmentation~\cite{hariharan2014simultaneous}. %

In this work, we ask if we may revisit the architecture design of wide residual networks to further boost the panoptic segmentation performance and even improve the model inference speed.
In particular, a baseline model is obtained by equipping the WR-41~\cite{chen2020naive} with the simple and effective modules, Squeeze-and-Excitation (SE)~\cite{hu2018squeeze} and Switchable Atrous Convoloution (SAC)~\cite{qiao2020detectors}. Similar to~\cite{wrn2016wide,howard2017mobilenets,tan2019efficientnet}, its network capacity could be further adjusted by three scaling factors $(w_1, w_2, \ell)$, where $w_1$ controls the channel size of the first two stages of a network backbone, while $w_2$ and $\ell$ adjust the network width (\ie, channel size) and depth (\ie, number of layers) of the remaining stages, respectively. The resulting model SWideRNet-$(w_1, w_2, \ell)$ (short for {\bf S}caling {\bf W}ide {\bf R}esidual Network with scaling factors $(w_1, w_2, \ell)$) is deployed as the network backbone in Panoptic-DeepLab~\cite{cheng2019panoptic} framework.

The SWideRNet-$(w_1, w_2, \ell)$ defines a large number of backbone architectures, targeting for different applications. The search space for those three scaling factors is discretized, allowing us to employ the simple and effective grid search method. Two model regimes are considered in this work, where the first one contains fast model variants and the other contains strong architectures. As a result, our main contribution lies in empirically identifying several fast SWideRNet backbones that attain state-of-the-art speed-accuracy trade-off, as well as several strong SWideRNet backbones that further push the envelope of panoptic segmentation benchmarks. As shown in \figref{fig:pq_vs_second}, our fast SWideRNets attain a better speed-accuracy trade-off than prior state-of-the-art models, showing at least 3\% PQ better than the MobileNetv3~\cite{howard2019searching} backbone at a similar speed. Interestingly, the found fast SWideRNets all share the same scaling factor $w_1=0.25$, indicating that the first two stages of Wide-ResNets are the speed bottleneck.
Finally, for the strong model regime, we found that going deeper (\ie, only increasing $\ell$) is the most efficient strategy to scale up the network capacity, suggesting that the Wide-ResNets may be already sufficiently `wide'. 
Our strong SWideRNet model variants consistently outperform prior {\it bottom-up} state-of-the-art Axial-DeepLab~\cite{wang2020axial} on three datasets.
Additionally, our {\it single} model outperforms ensemble models on Mapillary Vistas and ADE20K.

\section{Related Works}
\label{sec:related}

{\bf Convolutional Neural Networks:} 
Convolutional Neural Networks (CNNs)~\cite{lecun1998gradient} deployed in a fully convolutional manner (FCNs~\cite{sermanet2014overfeat,long2015fully}) have achieved remarkable performance on dense prediction tasks. The improvement of neural network design is one of the main driving forces for state-of-the-art systems, from AlexNet~\cite{krizhevsky2012imagenet}, VGG~\cite{simonyan2014very}, Inception~\cite{ioffe2015batch,szegedy2016rethinking,szegedy2017inception}, ResNet~\cite{he2016deep,he2016identity} to more recent architectures, such as  DenseNet~\cite{huang2017densely}, Xception~\cite{chollet2017xception, dai2017coco}, and EfficientNet~\cite{tan2019efficientnet}. Due to the simple yet effective design of residual networks~\cite{he2016deep}, there are many modern neural networks that build on top of it, including Wide-ResNet~\cite{wrn2016wide,wu2019wider}, ResNeXt~\cite{xie2017aggregated}, 
SENet~\cite{hu2018squeeze}, SKNet~\cite{li2019selective}
Res2Net~\cite{gao2019res2net}, RegNet~\cite{radosavovic2020designing}, and ResNeSt~\cite{zhang2020resnest}. 

\begin{figure*}[!t]
    \centering
    \scalebox{0.8}{
    \begin{tabular}{c | c}
    \includegraphics[height=0.46\textwidth]{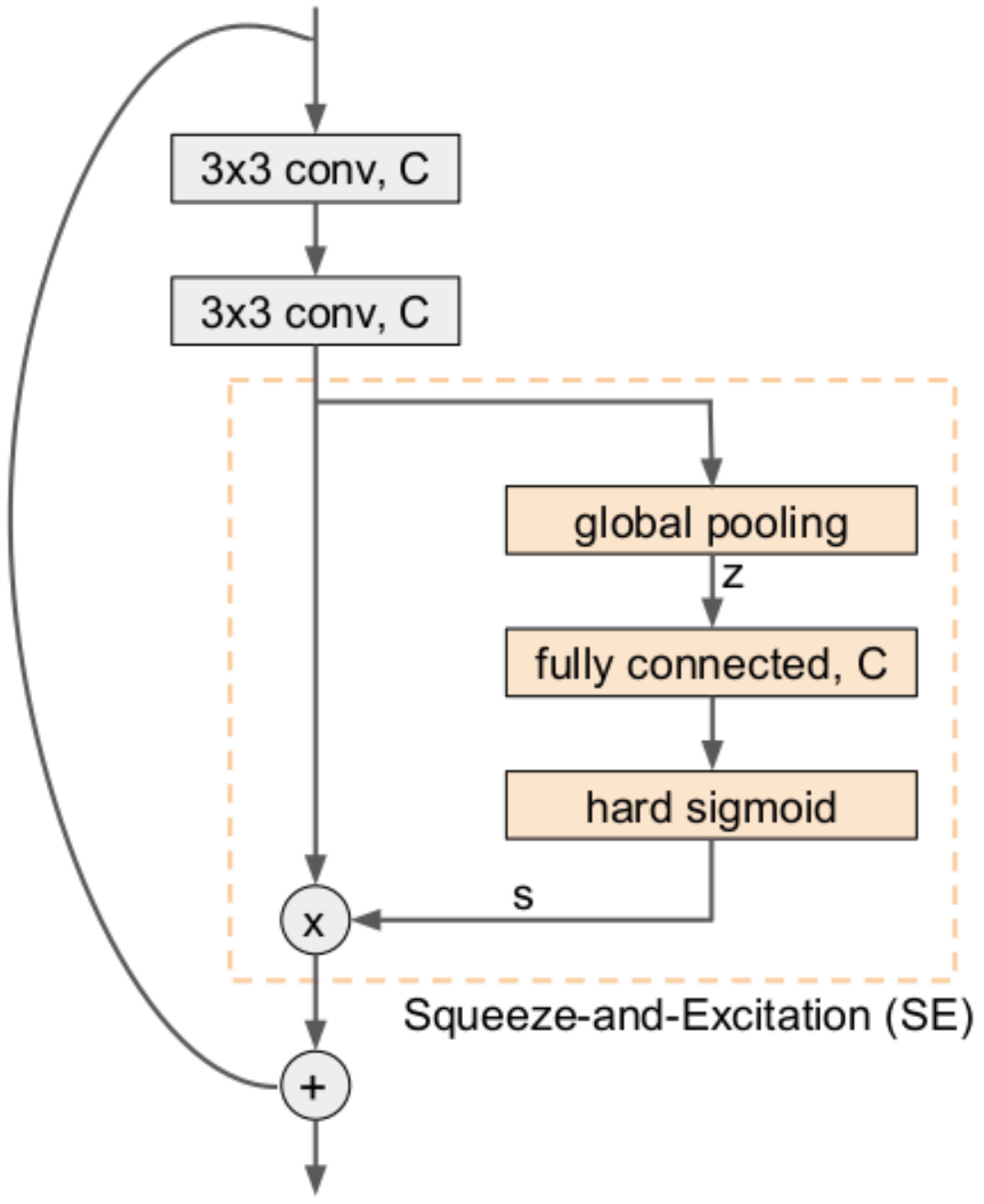} &
    \includegraphics[height=0.46\textwidth]{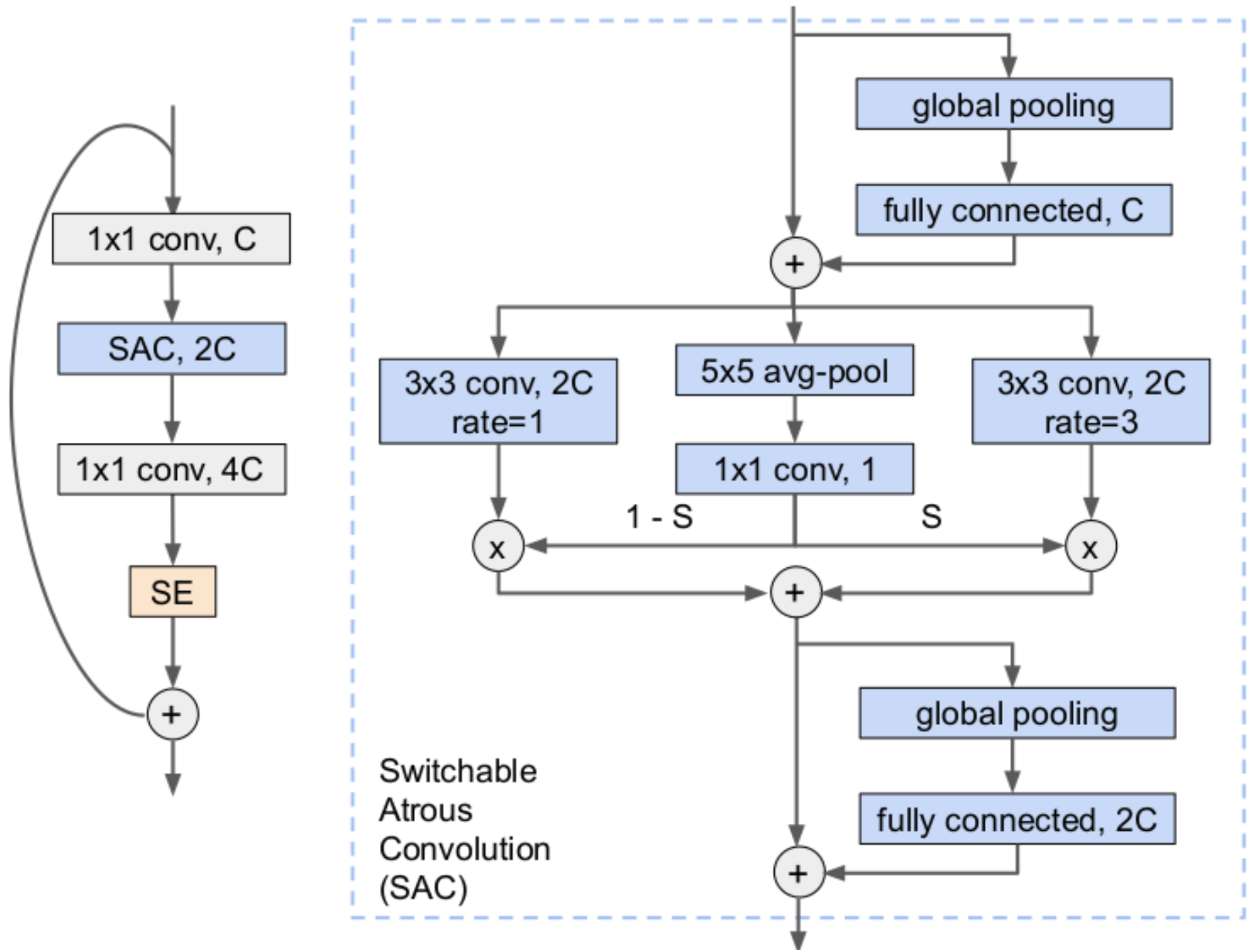} \\
    (a) Squeeze-and-Excitation &
    (b) Switchable Atrous Convolution \\
    \end{tabular}
    }
    \caption{Illustration of our employment of (a) Squeeze-and-Excitation (SE) module and (b) Swithcable Atrous Convolution (SAC).}
    \label{fig:modules}
\end{figure*}

{\bf Scaling CNNs:} The capacity of Convolutional Neural Networks (CNNs) could be scaled up by stacking more convolutional layers or increasing the channels. ResNet~\cite{he2016deep} is the first work that successfully stacks over 1000 convolutional layers for small-resolution images, while PSPNet~\cite{zhao2017pyramid} employs ResNet with 269 layers, and shows outstanding semantic segmentation results. MobileNets~\cite{howard2017mobilenets,sandler2018mobilenetv2,howard2019searching} and ShuffleNets~\cite{ma2018shufflenet,ma2018shufflenet} introduce a universal scaling factor to adjust network channels. Wide-ResNet~\cite{wrn2016wide,wu2019wider}, GPipe~\cite{gpipe18}, and BiT~\cite{kolesnikov2020big} explore scaling up both layers and channels for image classification. Auto-DeepLab~\cite{liu2019auto} increases the channels of a base network for better semantic segmentation performance. More recently, EfficientNet~\cite{tan2019efficientnet} and EfficientDet~\cite{tan2020efficientdet} adopt a compound factor to effectively and simultaneously scaling up layers, channels, and input resolutions for image classification and object detection, respectively. Our model follows the same direction by scaling the architecture of Wide-ResNet~\cite{wrn2016wide,wu2019wider}, specifically targeting for panoptic segmentation~\cite{kirillov2018panoptic}.

{\bf Panoptic Segmentation:} State-of-the-art panoptic segmentation systems could be roughly categorized into top-down (or proposal-based) and bottom-up (or box-free) approaches. Top-down approaches~\cite{kirillov2019panoptic,porzi2019seamless,li2018learning,li2018attention,liu2019e2e,xiong2019upsnet,li2020unifying,Chen_2020_CVPR,Lazarow_2020_CVPR,wu2020bidirectional,wu2020auto} typically pair Mask R-CNN~\cite{he2017mask} with a light-weight `stuff` segmentation branch, while bottom-up approaches~\cite{yang2019deeperlab,gao2019ssap,Wang_2020_CVPR,cheng2019panoptic,wang2020axial} group `thing` pixels from semantic segmentation predictions. Recently, Panoptic-DeepLab~\cite{cheng2019panoptic}, a simple yet effective bottom-up system for panoptic segmentation, employs DeepLab semantic segmentation outputs~\cite{chen2018deeplabv2,deeplabv3plus2018} coupled with a class-agnostic instance segmentation branch involving a simple instance center regression~\cite{kendall2018multi,uhrig2018box2pix,neven2019instance}. Panoptic-DeepLab~\cite{cheng2019panopticworkshop} has achieved state-of-the-art results on several benchmarks, and our method builds on top of it.

\section{Methods}
\label{sec:methods}
In this section, we describe how to effectively scale the capacity of our baseline model, obtained by incorporating to Wide-ResNet-41 (WR-41)~\cite{wrn2016wide,wu2019wider,chen2020naive} the simple yet effective Squeeze-and-Excitation~\cite{hu2018squeeze} and Switchable Atrous Convolution~\cite{qiao2020detectors} modules. The resulting network family with different scaling factors is then explored for both fast model regime and strong model regime.

\subsection{The SWideRNet family}
{\bf Baseline model:} The Wide Residual Networks~\cite{wrn2016wide,wu2019wider} have demonstrated outstanding performance on image classification~\cite{ILSVRC15}, object detection~\cite{lin2014microsoft}, and semantic segmentation~\cite{everingham2010pascal}. Specifically, the Wide-ResNet-38 (WR-38)~\cite{wu2019wider}, refined by several human-crafted networks with extensive experiments, has been the {\it de facto} network backbone for semantic segmentation~\cite{neuhold2017mapillary,rota2018place,zhu2019improving,takikawa2019gated,li2020improving}, and instance segmentation~\cite{liang2019polytransform,homayounfar2020levelset} on Cityscapes leader-board~\cite{cordts2016cityscapes}. Recently, Chen~\etal~\cite{chen2020naive} attain state-of-the-art panoptic segmentation performance~\cite{kirillov2018panoptic} on Cityscapes~\cite{cordts2016cityscapes} by employing the Wide-ResNet-41 (WR-41), which improves both accuracy and speed over the WR-38~\cite{wu2019wider} (when deploying in the Panoptic-DeepLab~\cite{cheng2019panoptic} framework) by (1) removing the last residual block, and (2) repeating the second last residual block two more times.

Building on top of WR-41, we further incorporate the simplified Squeeze-and-Excitation (SE) module~\cite{hu2018squeeze,lee2020centermask} (where only one fully connected layer is used), and the Switchable Atrous Convolution (SAC)~\cite{qiao2020detectors}, forming our baseline model. In  \figref{fig:modules}, we visualize the simplified SE module and SAC operation. To be concrete, the channel attention map $\mathbf{s}$ in the SE module is computed as follows.
\begin{equation}
    \mathbf{s} = \sigma(W\mathbf{z}),
\end{equation}
where $\mathbf{z}$ is the globally average-pooled input feature map, and $W$ are the weights of a fully connected layer. Following~\cite{howard2019searching}, we employ the hard sigmoid function~\cite{binaryconnect}: $\sigma(x) = \frac{ReLU6(x + 3)}{6}$.

\newcommand{\blocka}[2]{\multirow{3}{*}{\(\left[\begin{array}{c}\text{3$\times$3, #1}\\[-.1em] \text{3$\times$3, #1} \end{array}\right]\)$\times$#2}
}
\newcommand{\blockase}[2]{\multirow{3}{*}{\(\left[\begin{array}{c}\text{3$\times$3, #1}\\[-.1em] \text{3$\times$3, #1} \\ \text{SE}\end{array}\right]\)$\times$#2}
}

\newcommand{\blockb}[3]{\multirow{3}{*}{\(\left[\begin{array}{c}\text{3$\times$3, #1}\\[-.1em] \text{3$\times$3, #2} \end{array}\right]\)$\times$#3}
}
\newcommand{\blockc}[4]{\multirow{3}{*}{\(\left[\begin{array}{c}\text{1$\times$1, #1}\\[-.1em] \text{3$\times$3, #2}\\[-.1em] \text{1$\times$1, #3}\end{array}\right]\)$\times$#4}
}
\newcommand{\blockd}[4]{\multirow{4}{*}{\(\left[\begin{array}{c}\text{1$\times$1, #1}\\[-.1em] \text{3$\times$3 SAC, #2}\\[-.1em] \text{1$\times$1, #3}\\ \text{SE}\end{array}\right]\)$\times$#4}
}

\renewcommand\arraystretch{1.}
\setlength{\tabcolsep}{3pt}
\begin{table*}[!t]
\begin{center}
\resizebox{0.48\linewidth}{!}{
\begin{tabular}{c|c|c|c|c}
\hline
stage & input size & output size & WR-41 \cite{chen2020naive} & SWideRNet-($w_1$, $w_2$, $\ell$)\\
\hline
conv1 & 224$\times$224 & 112$\times$112 & 3$\times$3, 64, stride 2 & 3$\times$3, 64$\times w_1$, stride 2\\
\hline
\multirow{4}{*}{conv2}  & \multirow{4}{*}{112$\times$112}  & \multirow{4}{*}{56$\times$56} & \blocka{128}{3}  & \blocka{128$\times w_1$}{3}\\
  &  &  &  \\
  &  &  &  \\
  \cline{4-5}
  &  &  & \multicolumn{2}{c}{3$\times$3 max-pool, stride 2} \\
\hline
\multirow{5}{*}{conv3}  & \multirow{5}{*}{56$\times$56}  & \multirow{5}{*}{28$\times$28}  &   & \\
  &  &  &  \blocka{256}{3} & \blockase{256$\times w_2$}{3$\ell$} \\
  &  &  &  & \\
  &  &  &  & \\
  &  &  &  & \\
\hline
\multirow{5}{*}{conv4}  & \multirow{5}{*}{28$\times$28}  & \multirow{5}{*}{14$\times$14}  &   & \\
  &  &  &  \blocka{512}{6} & \blockase{512$\times w_2$}{6$\ell$}\\
  &  &  &  & \\
  &  &  &  & \\
  &  &  &  & \\
\hline
\multirow{5}{*}{conv5}  & \multirow{5}{*}{14$\times$14}  & \multirow{5}{*}{7$\times$7}  &   & \\
  &  &  &  \blocka{1024}{3} & \blockase{1024$\times w_2$}{3$\ell$} \\
  &  &  &  & \\
  &  &  &  & \\
  &  &  &  & \\
\hline
\multirow{6}{*}{conv6}  & \multirow{6}{*}{7$\times$7}  & \multirow{6}{*}{7$\times$7}  &   & \\
  &  &  & \blockc{512}{1024}{2048}{3}   & \blockd{512$\times w_2$}{1024$\times w_2$}{2048$\times w_2$}{3$\ell$}\\
  &  &  &  & \\
  &  &  &  & \\
  &  &  &  & \\
  &  &  &  & \\
\hline

& 7$\times$7 & 1$\times$1  & \multicolumn{2}{c}{average-pool, 1000-d fc, softmax} \\
\hline

\end{tabular}
}
\end{center}
\caption{Architectures for our implementation of Wide-ResNet-41 (WR-41) and our  SWideRNet-($w_1$, $w_2$, $\ell$) on ImageNet. SE denotes the simplified squeeze-and-excitation module (where only one fully connected layer is used) and SAC denotes the Switchable Atrous Convolution. Our SWideRNet scales the channels and layers of WR-41 to obtain a family of network backbones for panoptic segmentation. Note, if not specified, a strided convolution is used in the last residual block if spatial resolution changes.
}
\label{tab:arch}
\end{table*}

The SAC operation essentially gathers features computed with different atrous rates~\cite{holschneider1989real,papandreou2014untangling,deeplabv12015}. Specifically, we use $\mathbf{y}=\text{\bf Conv}(\mathbf{x}, \mathbf{w}, r)$ to  denote the convolutional operation with weights $\mathbf{w}$, atrous rate $r$, input $\mathbf{x}$, and output $\mathbf{y}$. SAC adopts a switch function $\mathbf{S}$ to merge two feature maps:
\begin{equation}
    (1 - \mathbf{S}(\mathbf{x})) \cdot \text{\bf Conv}(\mathbf{x}, \mathbf{w}, 1) + \mathbf{S}(\mathbf{x}) \cdot \text{\bf Conv}(\mathbf{x}, \mathbf{w}, 3),
\end{equation}
where we use $r$ = 1 and 3 for two convolutions (with weights $\mathbf{w}$ shared). The switch function $\mathbf{S}$ is input- and location-dependent. It  is implemented as a $5\times5$ average pooling followed by a $1\times1$ convolution. Following~\cite{qiao2020detectors}, we also insert two global context modules before and after the main operation of SAC. Those global context modules are lightweight and are implemented as global average pooling followed by a fully connected layer. We use ordinary convolution in the SAC operation (\ie, no deformation~\cite{dai2017deformable}). 

{\bf Scaling factors:} Similar to~\cite{wrn2016wide,howard2017mobilenets,tan2019efficientnet}, we adopt scaling factors, $(w_1, w_2, \ell)$, to scale the network capacity of our baseline model, where $w_1$ scales the channels of the first two stages (denoted as conv1 and conv2), $w_2$ and $\ell$ scale the channels and layers of the remaining stages (denoted as conv3, conv4, conv5, and conv6), respectively. The resulting family of networks is dubbed SWideRNet-$(w_1, w_2, \ell)$ for {\bf S}caling {\bf W}ide {\bf R}esidual Networks with scaling factors $(w_1, w_2, \ell)$. We illustrate the network architecture in \tabref{tab:arch}. The total number of layers in the network backbone is thus equal to $7 + 33 \times \ell$. Note that this calculation does not include the SE and extra operations incurred by the SAC.

\subsection{Exploring SWideRNet}

The SWideRNet-$(w_1, w_2, \ell)$ family defines abundant network architectures. One thus could search for different SWideRNet architectures, designed for different objectives and applications. In this work, we apply SWideRNets to panoptic segmentation~\cite{kirillov2018panoptic} for two scenarios. In the first scenario, we target at designing fast SWideRNets that attain  state-of-the-art speed-accuracy trade-off (\ie, PQ \vs GPU runtime), which is applicable to on-device panoptic segmentation. The latency speed is directly measured by a GPU, instead of by any proxy (\eg, M-Adds). In the second scenario, we aim for state-of-the-art accuracy regardless of some costs (\eg, model parameters and speed), which could be deployed in cloud or server-side panoptic segmentation. 

{\bf Grid search: } The search space of SWideRNet-$(w_1, w_2, \ell)$ is discretized, allowing us to employ the simple yet effective grid search method. We elaborate on the discretized search space for each scenario below.

{\bf Fast mode regime: } We constrain SWideRNet-$(w_1, w_2, \ell)$ to be in the search space $\boldsymbol{S}_{fast}$ by scaling down the network capacity for fast inference speed, resulting in a total of only 45 architecture candidates.
\begin{align}\label{eq:space_fast}
    \boldsymbol{S}_{fast} = \{(w_1, w_2, \ell) | & w_1 \in \{0.25, 0.5, 1\}, \nonumber \\
    & w_2 \in \{0.25, 0.35, 0.5, 0.75, 1\}, \nonumber\\
    & \ell \in \{0.35, 0.75, 1\} \},
\end{align}

{\bf Strong model regime: } We scale up the network capacity for better prediction accuracy by considering SWideRNet-$(w_1, w_2, \ell)$ in the search space $\boldsymbol{S}_{strong}$, resulting in a total of 21 architecture candidates. In practice, we only experiment with 11 candidates since not all the candidates could fit into the GPU/TPU memories.
\begin{align}\label{eq:space_strong}
    \boldsymbol{S}_{strong} = \{(w_1, w_2, \ell) | & w_1 \in \{1\}, \nonumber \\
    & w_2 \in \{1, 1.5, 2\}, \nonumber\\
    & \ell \in \{1, 2, 3, 4, 5, 5.5, 6\} \},
\end{align}

\section{Experimental Results}
\label{sec:experiments}

We conduct experiments on several datasets.

{\bf COCO \cite{lin2014microsoft}:} There are 118K, 5K, and 20K images for training, validation, and testing, respectively. The dataset consists of 80 `thing` and 53 `stuff` classes.

{\bf Cityscapes \cite{cordts2016cityscapes}:} The dataset consists of 2975, 500, and 1525 traffic-related images for training, validation, and testing, respectively. It contains 8 `thing' and 11 `stuff' classes.

{\bf Mapillary Vistas \cite{neuhold2017mapillary}:} A large-scale traffic-related dataset, containing 18K, 2K, and 5K images for training, validation and testing, respectively. It contains 37 `thing' classes and 28 `stuff' classes in a variety of image resolutions, ranging from $1024\times768$ to more than $4000\times6000$

{\bf ADE20K \cite{zhou2019semantic}:} A high-quality densely annotated dataset, consisting of 20K, 2K, and 3K images for training, validation and testing, respectively. There are 100 `thing` and 50 `stuff` classes.

{\bf Experimental setup:} We report mean IoU, average precision (AP), and panoptic quality (PQ) to evaluate the semantic, instance, and panoptic segmentation results.

Our proposed SWideRNet is employed as the backbone in  Panoptic-DeepLab~\cite{cheng2019panoptic}. We follow closely the experimental setup of \cite{cheng2019panoptic}. For example, all our models are trained using TensorFlow~\cite{tensorflow-osdi2016} on 32 TPUs with the `poly' learning rate policy~\cite{liu2015parsenet} and an initial learning rate of $0.0001$. We  fine-tune the batch normalization~\cite{ioffe2015batch} 
parameters, perform random scale data augmentation during training, and optimize with Adam \cite{kingma2015adam} without weight decay.

On COCO, our models are trained with crop size $641\times641$ and batch size 64. Few models are trained with crop size $1025\times1025$ for better accuracy and we will clearly specify it.
On Cityscapes, we use crop size $1025\times2049$ and batch size 32, while on Mapillary Vistas, the images are resized to 2049 pixels at the longest side, and we randomly crop $1025\times2049$ patches during training with batch size 32.
Finally, our model is trained with crop size $641\times641$ with batch size 64 on ADE20K.

We set training iterations to 500K, 60K, 500K, 180K for COCO, Cityscapes, Mapillary Vistas, and ADE20K, respectively. We employ the same loss functions and loss weights as Panoptic-DeepLab~\cite{cheng2019panoptic}. When training some of our large model variants, we adopt AutoAugment~\cite{cubuk2018autoaugment} with the augmentation policy defined in \tabref{tab:autoaugment}.

During evaluation, due to the sensitivity of PQ \cite{xiong2019upsnet,li2018learning,porzi2019seamless}, we re-assign to `VOID' label all `stuff' segments whose areas are smaller than a threshold. The thresholds on COCO, Cityscapes, Mapillary Vistas, and ADE20K are 4096, 2048, 4096, and 4096, respectively. Additionally, we adopt multi-scale inference (scales equal to $\{0.5, 0.75, 1, 1.25, 1.5\}$ for COCO and ADE20K, and $\{0.5, 0.75, 1, 1.25, 1.5, 1.75, 2\}$ for Cityscapes and Mapillary Vistas) and left-right flipped inputs.

\begin{table}[!t]
  \centering
  \scalebox{0.8}{
  \begin{tabular}{c | c  c  c | c  c  c }
    \toprule[0.2em]
    & {\bf Op 1} & {\bf Prob} & {\bf Mag} & {\bf Op 2} & {\bf Prob} & {\bf Mag} \\
    \toprule[0.2em]
    Sub-policy 1 & Sharpness & 0.4 & 1.4 & Brightness & 0.2 & 2.0 \\
    Sub-policy 2 & Equalize & 0 & 1.8 & Contrast & 0.2 & 2.0 \\
    Sub-policy 3 & Sharpness & 0.2 & 1.8 & Color & 0.2 & 1.8 \\
    Sub-policy 4 & Solarize & 0.2 & 1.4 & Equalize & 0.6 & 1.8 \\
    Sub-policy 5 & Sharpness & 0.2 & 0.2 & Equalize & 0.2 & 1.4 \\
    \bottomrule[0.1em]
  \end{tabular}
  }
  \caption{Augmentation policy used in our experiments. We refer readers to AutoAugment \cite{cubuk2018autoaugment} for details of augmentation operations. Each sub-policy consists of two {\bf Op}erations with different {\bf Prob}abilities and {\bf Mag}nitudes. During training, one of the sub-policies is selected uniformly at random.}
  \label{tab:autoaugment}
  \vspace{-2mm}
\end{table}

\subsection{Ablation Studies}

We perform ablation studies on the validation set of COCO panoptic segmentation.

{\bf Design choices:} Our system builds on top of Panoptic-DeepLab~\cite{cheng2019panoptic} by deploying different backbone architectures. In \tabref{tab:coco_design_choices}, we report the effect of  incorporating new modules to our baseline, Wide-ResNet-41 (WR-41)~\cite{chen2020naive}. Adopting the multi-grid scheme \cite{chen2017deeplabv3,wang2017understanding} in the last three residual blocks (with unit rate $\{1, 2, 4\}$, same as \cite{chen2017deeplabv3}) improves the performance by 0.5\% PQ with extra marginal computation overhead (but no extra parameters). However, it is more effective to employ the Switchable Atrous Convolution (SAC)~\cite{qiao2020detectors} in the last three residual blocks, which improves over the baseline by 1.2\% PQ with small computational overhead. Additionally, adding Squeeze-and-Excitation (SE) modules could further improve the performance by 0.6\% PQ. Finally, we notice that employing separable convolutions~\cite{howard2017mobilenets} in the ASPP and decoder modules, same as the original design of Panoptic-DeepLab~\cite{cheng2019panoptic}, only degrades the performance by 0.2\% PQ while the inference speed is significantly improved. Therefore, for the fast model regime, we employ separable convolutions in the ASPP and decoder modules, while original convolutions are used for the strong model regime.

{\bf Training tricks:} During training, we employ drop path~\cite{huang2016deep} with a constant survival rate 0.8, and AutoAugment~\cite{cubuk2018autoaugment} with policy defined in \tabref{tab:autoaugment}, which improve 0.2\% PQ, and 0.3\% PQ, respectively. Note that in all the reported experimental results, we only apply AutoAugment to the strongest model for test server evaluation.

\begin{figure}
    \centering
    \scalebox{1}{
    \begin{tabular}{c c}
    \includegraphics[height=0.12\textwidth]{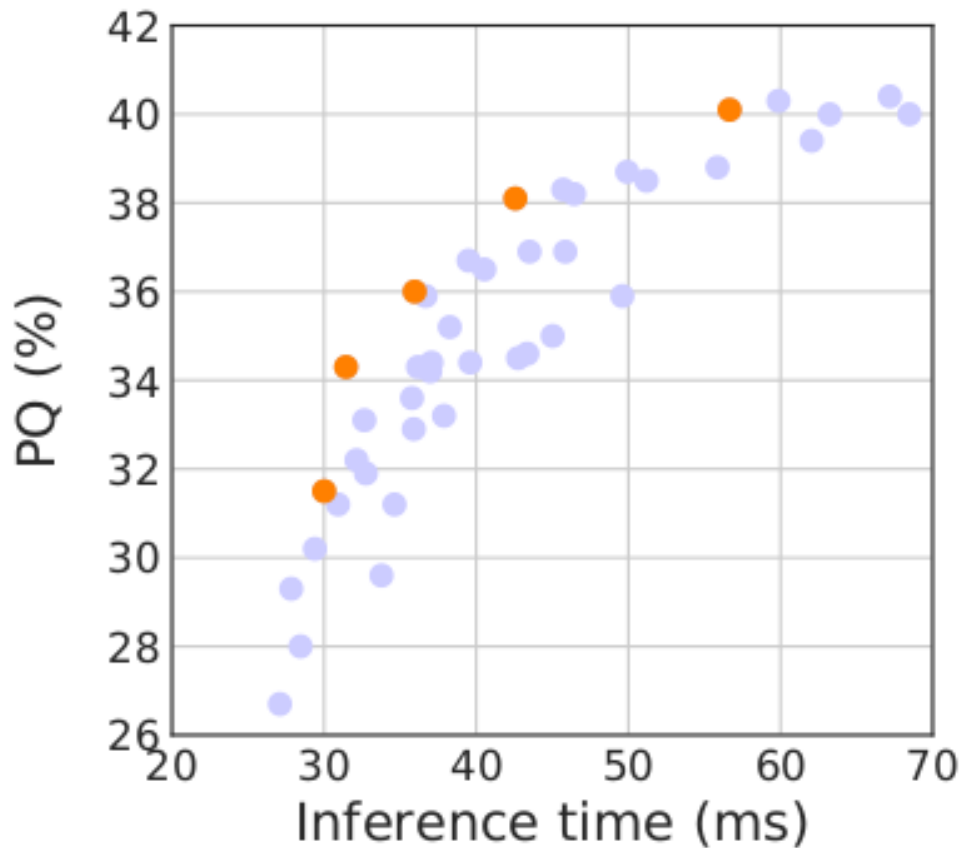} &
    \includegraphics[height=0.12\textwidth]{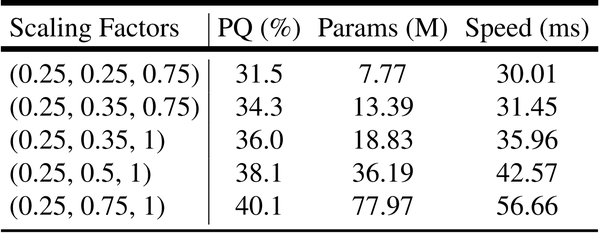} \\
    (a) & (b) \\
    \end{tabular}
    }
    \caption{(a) Grid search for fast model variants. (b) Five selected fast SWideRNets (marked in orange in (a)).}
    \label{fig:grid_search_fast}
\end{figure}

{\bf Fast model regime:} We conduct grid search in the fast model regime with search space defined by Eq.~\eqref{eq:space_fast} where the channels and layers are scaled {\it down} for faster inference. \figref{fig:grid_search_fast}~(a) shows the scatter plot of PQ \vs GPU inference time (Tesla V100-SXM2). We pinpoint five candidate architectures (marked in orange) that attain the best speed-accuracy trade-off, as shown in~\figref{fig:grid_search_fast}~(b). The found fast architectures share the same scaling factor $w_1=0.25$, indicating that the stages conv1 and conv2 are the speed bottleneck. We report detailed comparison with other state-of-the-art models in the following \secref{sec:runtime}.

{\bf Strong model regime:} Similarly, we perform grid search in the strong model regime with search space defined by Eq.~\eqref{eq:space_strong} where the channels and layers are scaled {\it up} for better accuracy. As shown in \tabref{tab:grid_search_large_models}, we find that `Going Deeper' (\ie, increasing only layers $\ell$) is more efficient than both `Going Wider' (\ie, increasing only channels $w_2$) and `Going Wider and Deeper' (\ie, increasing both channels $w_2$ and layers $\ell$). The finding suggests that the Wide-ResNets may be already sufficiently `wide' on current benchmarks.
The SWideRNet-(1, 1, 5.5) attains the best accuracy with crop size $641\times641$. When the crop size is increased to $1025\times1025$, the SWideRNet-(1, 1, 4) further improves the performance to 45.3\% PQ, but with a slower inference speed. This backbone is selected for COCO test-dev evaluation.
Finally, we adopt the same `Going Deeper' strategy on the other datasets for strong model regime.

\begin{table}[!t]
  \centering
  \scalebox{0.76}{
  \begin{tabular}{c c c c | c c c c }
    \toprule[0.2em]
    MG & SAC & SE & Sep-Conv & PQ (\%) & Params (M) & M-Adds (B) & Runtime (ms)\\
    \toprule[0.2em]
           &        &        &        & 39.6 & 147.32 & 655.66 & 101.16 \\
    \cmark &        &        &        & 40.1 & 147.32 & 659.02 & 104.39 \\
           & \cmark &        &        & 40.8 & 151.26 & 680.68 & 105.91 \\
           & \cmark & \cmark &        & 41.4 & 168.77 & 680.79 & 108.36 \\
           & \cmark & \cmark & \cmark & 41.2 & 136.92 & 499.16 & 88.48 \\
    \bottomrule[0.1em]
  \end{tabular}
  }
  \caption{Design choices on COCO {\it val} set. The baseline corresponds to WR-41. {\bf MG:} Multi-Grid. {\bf SAC:} Switchable Atrous Convolution. {\bf SE:} Squeeze-and-Excitation. {\bf Sep-Conv:} Employing separable convolutions in ASPP and decoder modules.
  }
  \label{tab:coco_design_choices}
  \vspace{-2mm}
\end{table}

\begin{table}[!t]
  \centering
  \scalebox{0.76}{
  \begin{tabular}{l | c c c c }
    \toprule[0.2em]
    Backbone & PQ (\%) & Params (M) & M-Adds (B) & Runtime (ms)\\
    \toprule[0.2em]
    \multicolumn{5}{c}{Baseline Model}\\
    \midrule
    SWideRNet-(1, 1, 1) & 41.4 & 168.77 & 680.79 & 108.36 \\
    \midrule
    \multicolumn{5}{c}{Going Wider}\\
    \midrule
    SWideRNet-(1, 1.5, 1) & 42.8 & 345.75 & 1205.95 & 175.84 \\
    SWideRNet-(1, 2, 1)   & 43.0 & 587.35 & 1925.75 & 266.85 \\
    \midrule
    \multicolumn{5}{c}{Going Deeper}\\
    \midrule
    SWideRNet-(1, 1, 2) & 42.8 & 302.33 & 1119.77 & 190.37 \\
    SWideRNet-(1, 1, 3) & 43.3 & 435.90 & 1558.75 & 310.90 \\
    SWideRNet-(1, 1, 4) & 43.8 & 569.46 & 1997.73 & 407.88 \\
    SWideRNet-(1, 1, 5) & 44.0 & 703.03 & 2436.71 & 452.42 \\
    SWideRNet-(1, 1, 5.5) & 44.1 & 752.53 & 2614.01 & 504.66 \\
    SWideRNet-(1, 1, 5.5)$\dagger$ & 44.4 & 752.53 & 2614.01 & 504.66 \\
    SWideRNet-(1, 1, 6) & 44.0 & 836.59 & 2875.69 & 544.02 \\
    \midrule
    \multicolumn{5}{c}{Going Wider and Deeper}\\
    \midrule
    SWideRNet-(1, 1.5, 2) & 43.7 & 646.22 & 2193.54 & 344.74 \\
    SWideRNet-(1, 1.5, 3) & 44.1 & 946.69 & 3181.12 & 570.73 \\
    \midrule
    \multicolumn{5}{c}{Going Deeper with Large Crop Size $1025\times1025$}\\
    \midrule
    SWideRNet-(1, 1, 3)$\dagger$ & 44.7 & 435.90 & 3911.58 & 641.21 \\
    SWideRNet-(1, 1, 4)$\dagger$ & 45.3 & 569.46 & 5019.62 & 836.41 \\
    \bottomrule[0.1em]
  \end{tabular}
  }
  \caption{Grid search for large models.
  $\dagger:$ Use AutoAugment.
  }
  \label{tab:grid_search_large_models}
  \vspace{-2mm}
\end{table}

\subsection{Fast Model Regime}
\label{sec:runtime}
In \tabref{tab:runtime}, we compare our five fast SWideRNet model variants with other state-of-the-art models on both COCO and Cityscapes. We report the end-to-end runtime (\ie, inference time from an input image to final panoptic segmentation result, including {\it all} operations such as merging semantic and instance segmentation). The inference speed is measured on a Tesla V100-SXM2 GPU \emph{with batch size of one}. Additionally, \figref{fig:pq_vs_second} shows the scatter plot of speed \vs accuracy. As shown in the table and figure, our models attain better speed-accuracy trade-off than all state-of-the-art models. Specifically, on the COCO dataset, employing our SWideRNet-(0.25, 0.35, 1) is 6\% and 6.4\% PQ better than using MobileNetv3~\cite{howard2019searching} as backbone in Panoptic-DeepLab~\cite{cheng2019panoptic} on {\it val} and {\it test-dev} set, respectively, while a similar inference speed is achieved. Using SWideRNet-(0.25, 0.5, 1) is 3\% PQ better than ResNet-50~\cite{he2016deep} on both {\it val} and {\it test-dev} sets, while our model is slightly faster. Finally, adopting our SWideRNet-(0.25, 0.75, 1) achieves a similar performance to Xception-71~\cite{chollet2017xception,dai2017coco} (with $1025\times1025$ input) but with 2.3 times faster inference speed. On the Cityscapes dataset, when considering a similar inference speed, using our SWideRNet-(0.25, 0.25, 0.75) is 3\% and 2.5\% PQ better than using MobileNetv3 on {\it val} and {\it test} set, respectively. Employing SWideRNet-(0.25, 0.5, 1) is also 3\% and 2.8\% PQ better than ResNet-50 on {\it val} and {\it test} set, respectively. Our SWideRNet-(0.25, 0.75, 1) is also slightly faster and better than  Xception-71.

\begin{table*}[!t]
  \centering
  \scalebox{0.77}{
  \begin{tabular}{l | l | c | c  c  c | c  c  c | c  c }
    \toprule[0.2em]
    Method & Backbone & Input Size & PQ [val] & AP [val] & mIoU [val] & PQ [test] & AP [test] & mIoU [test] & Speed (ms) & M-Adds (B) \\
    \toprule[0.2em]
    Panoptic-DeepLab & SWideRNet-(0.25, 0.25, 0.75) & $1025\times2049$ & 58.4 & 30.2 & 77.6 & 56.6 & 25.5 & 76.9 & 63.05 & 151.97 \\
    Panoptic-DeepLab & SWideRNet-(0.25, 0.35, 0.75) & $1025\times2049$ & 59.8 & 31.3 & 79.4 & 58.2 & 27.3 & 78.6 & 73.14 & 229.49 \\
    Panoptic-DeepLab & SWideRNet-(0.25, 0.35, 1) & $1025\times2049$ & 61.6 & 33.3 & 80.2 & 60.0  & 29.6 & 79.6 & 83.25 & 319.14 \\
    Panoptic-DeepLab & SWideRNet-(0.25, 0.5, 1) & $1025\times2049$ & 62.7 & 35.2 & 80.3 & 60.8 & 30.5 & 79.8 & 108.58 & 576.83 \\
    Panoptic-DeepLab & SWideRNet-(0.25, 0.75, 1) & $1025\times2049$ & 63.8 & 36.1 & 80.6 & 61.6 & 31.2 & 80.6 & 166.83 & 1199.85 \\
    \bottomrule[0.1em]
  \end{tabular}
  }
  \caption{Cityscapes end-to-end runtime results, including merging semantic and instance segmentation. All numbers are obtained by (1) a single-scale input (batch size {\it one}) without flipping, and (2) built-in TensorFlow library without extra inference optimization. {[val]:} Performance on val set. {[test]:} Performance on test  set. We report PQ, AP, and mIoU for panoptic, instance, and semantic segmentation.
  }
  \label{tab:runtime_sup}
\end{table*}

{\bf More fast model results on Cityscapes:}
Our methods build on top of Panoptic-DeepLab~\cite{cheng2019panoptic}, simultaneously generating semantic segmentation, instance segmentation, and panoptic segmentation results (\ie, no need for specific task fine-tuning). Therefore, in \tabref{tab:runtime_sup}, we additionally report the instance segmentation and semantic segmentation results for our fast model variants, evaluated on both Cityscapes validation and test sets~\cite{cordts2016cityscapes}.

\subsection{Strong Model Regime}
For the strong model regime, we adopt the "Going Deeper" strategy (\ie, only scaling up the number of layers). We report the results for each dataset below.

{\bf COCO:} \tabref{tab:coco_val} summarizes our {\it val} set results. Our SWideRNet-(1, 1, 4) without multi-scale inference already outperforms Aixal-DeepLab~\cite{wang2020axial}. Incorporating the multi-scale inference further improves the performance to 45.8\% PQ. In \tabref{tab:coco_test}, our SWideRNet-(1, 1, 4) achieves 46.7\% PQ on {\it test-dev}, outperforming the current {\it bottom-up} state-of-the-art Axial-DeepLab-L~\cite{wang2020axial} by 2.5\% PQ. Note that for this result, we increase the magnitudes of employed AutoAugment policy (\tabref{tab:autoaugment}) by a factor of 5, which additionally improves the performance by 0.2\% PQ.

{\bf Cityscapes:} As shown in \tabref{tab:cityscapes_val}, when using multi-scale inference, our SWideRNet-(1, 1, 4.5) outperforms Axial-DeepLab-XL~\cite{wang2020axial} by 2.4\% PQ (4.9\% AP) with only Cityscapes fine annotations, and 1.1\% PQ (2.6\% AP) with extra Mapillary Vistas pretraining~\cite{neuhold2017mapillary}. We then report the test set results in \tabref{tab:cityscapes_test}. When only using Cityscapes fine annotations, our model significantly outperforms current state-of-the-art Axial-DeepLab-XL~\cite{wang2020axial} by 2\% PQ and 4\% AP. When using extra data~\cite{neuhold2017mapillary}, our model achieves 67.8\% PQ, 42.2\% AP, and 84.1\% mIOU, better than Axial-DeepLab-XL~\cite{wang2020axial} by 1.2\% PQ and 2.6\% AP.
Following Naive-Student~\cite{chen2020naive}, with additional pseudo-labels~\cite{lee2013pseudo,zhu2019improving,xie2019self} from Cityscapes video and train-extra sets, our model reaches the performance of 68.5\% PQ and 43.4\% AP, setting a new state-of-the-art.

{\bf Mapillary Vistas:} \tabref{tab:mv_val} summarizes our {\it val} set results. Our SWideRNet-(1, 1, 4.5), with multi-scale inference, attains 44.8\% PQ, 22.2\% AP, and 60.0\% mIoU, outperforming the {\it bottom-up} (or box-free) method Axial-DeepLab-L~\cite{wang2020axial} by 3.7\% PQ, 5.0\% AP, and 1.6\% mIoU. Remarkably, our {\it single} model even performs better than the ensemble of six Panoptic-DeepLab models~\cite{cheng2019panoptic} by 2.6\% PQ, 4.0\% AP, and 1.3\% mIoU.

{\bf ADE20K:} In \tabref{tab:ade20k_val}, we report our results on ADE20K. On the {\it val} set, our SWideRNet-(1, 1, 4) significantly outperforms BGRNet~\cite{wu2020bidirectional} by 6\% PQ and Auto-Panoptic~\cite{wu2020auto} by 5.4\% PQ. Additionally, our model achieves 49.96\% mIoU and 83.78\% Pixel-accuracy. On the test set, our {\it single} model yields the score of 59.14\% (average of mIoU and Pixel-accuracy), 1.9\% better than the ensemble of PSPNets~\cite{zhao2017pyramid}, setting a new state-of-the-art. Interestingly, our SWideRNet-(1, 1.5, 3) achieves a higher score of 50.35\% mIoU on the validation set, but a lower score of 40.47\% mIoU on the test set, presenting another challenge in ADE20K to avoid over-fitting large models.

\begin{table}[!t]
  \centering
  \scalebox{0.5}{
  \begin{tabular}{l | l | c | c | c | c | c }
    \toprule[0.2em]
    Method & Backbone & Input Size & PQ [val] & PQ [test] & Speed (ms) & M-Adds (B) \\
    \toprule[0.2em]
    \multicolumn{7}{c}{COCO}\\
    \midrule
    DeeperLab~\cite{yang2019deeperlab} & W-MNV2 & $641\times641$ & 27.9 & 28.1 & 83 & - \\
    DeeperLab~\cite{yang2019deeperlab} & Xception-71 & $641\times641$ & 33.8 & 34.3 & 119 & - \\
    \midrule
    Hou~\etal~\cite{hou2020real} & ResNet-50 & $800\times1333$ & 37.1 & - & 63 & - \\
    \midrule
    PCV~\cite{Wang_2020_CVPR} & ResNet-50 & $800\times1333$ & 37.5 & 37.7 & 176.5 & - \\
    \midrule
    Panoptic-DeepLab~\cite{cheng2019panoptic} & MobileNetv3 & $641\times641$ & 30.0 & 29.8 & 38 & 12.24 \\
    Panoptic-DeepLab~\cite{cheng2019panoptic} & ResNet-50 & $641\times641$ & 35.1 & 35.2 & 50 & 77.79 \\
    Panoptic-DeepLab~\cite{cheng2019panoptic} & Xception-71 & $641\times641$ & 38.9 & 38.8 & 74 & 109.21 \\
    Panoptic-DeepLab~\cite{cheng2019panoptic} & Xception-71 & $1025\times1025$ & 39.7 & 39.6 & 132 & 279.25 \\
    \midrule
    UPSNet~\cite{xiong2019upsnet} & ResNet-50 & $800\times1333$ & 42.5 & - & 167 & - \\
    \midrule \midrule
    Panoptic-DeepLab & SWideRNet-(0.25, 0.25, 0.75) & $641\times641$ & 31.5 & 31.7 & 30.01 & 31.67 \\
    Panoptic-DeepLab & SWideRNet-(0.25, 0.35, 0.75) & $641\times641$ & 34.3 & 34.4 & 31.45 & 47.13 \\
    Panoptic-DeepLab & SWideRNet-(0.25, 0.35, 1) & $641\times641$ & 36.0 & 36.2 & 35.96 & 64.98 \\
    Panoptic-DeepLab & SWideRNet-(0.25, 0.5, 1) & $641\times641$ & 38.1 & 38.2 & 42.57 & 116.37 \\
    Panoptic-DeepLab & SWideRNet-(0.25, 0.75, 1) & $641\times641$ & 40.1 & 40.3 & 56.66 & 240.59 \\
    \midrule
    \multicolumn{7}{c}{Cityscapes}\\
    \midrule
    PCV~\cite{Wang_2020_CVPR} & ResNet-50 & $1024\times2048$ & 54.2 & - & 182.8 & - \\
    \midrule
    DeeperLab~\cite{yang2019deeperlab} & W-MNV2 \cite{sandler2018mobilenetv2} & $1025\times2049$ & 52.3 & - & 303 & - \\
    DeeperLab~\cite{yang2019deeperlab} & Xception-71 & $1025\times2049$ & 56.5 & - & 463 & - \\
    \midrule
    Hou~\etal~\cite{hou2020real} & ResNet-50 & $1024\times2048$ & 58.8 & - & 99 & - \\
    \midrule
    UPSNet~\cite{xiong2019upsnet} & ResNet-50 & $1024\times2048$ & 59.3 & - & 202 & - \\
    \midrule
    Panoptic-DeepLab~\cite{cheng2019panoptic} & MobileNetv3 & $1025\times2049$ & 55.4 & 54.1 & 63 & 54.17 \\
    Panoptic-DeepLab~\cite{cheng2019panoptic} & ResNet-50 & $1025\times2049$ & 59.7 & 58.0 & 117 & 381.39 \\
    Panoptic-DeepLab~\cite{cheng2019panoptic} & Xception-71 & $1025\times2049$ & 63.0 & 60.7 & 175 & 547.49 \\
    \midrule \midrule
    Panoptic-DeepLab & SWideRNet-(0.25, 0.25, 0.75) & $1025\times2049$ & 58.4 & 56.6 & 63.05 & 151.97 \\
    Panoptic-DeepLab & SWideRNet-(0.25, 0.35, 0.75) & $1025\times2049$ & 59.8 & 58.2 & 73.14 & 229.49 \\
    Panoptic-DeepLab & SWideRNet-(0.25, 0.35, 1) & $1025\times2049$ & 61.6 & 60
    0 & 83.25 & 319.14 \\
    Panoptic-DeepLab & SWideRNet-(0.25, 0.5, 1) & $1025\times2049$ & 62.7 & 60.8 & 108.58 & 576.83 \\
    Panoptic-DeepLab & SWideRNet-(0.25, 0.75, 1) & $1025\times2049$ & 63.8 & 61.6 & 166.83 & 1199.85 \\
    \bottomrule[0.1em]
  \end{tabular}
  }
  \caption{End-to-end runtime, including merging semantic and instance segmentation. All results are obtained by (1) a single-scale input without flipping, and (2) built-in TensorFlow library without extra inference optimization. {\bf PQ [val]:} PQ (\%) on val set. {\bf PQ [test]:} PQ (\%) on test(-dev) set.
  }
  \label{tab:runtime}
\end{table}

\begin{table}[!t]
  \centering
  \scalebox{0.7}{
  \begin{tabular}{c | c | c | c | c | c }
    \toprule[0.2em]
    Method & Backbone & MS  & PQ (\%) & $\text{PQ}^{\text{Th}}$ (\%) & $\text{PQ}^{\text{St}}$ (\%)\\
    \toprule[0.2em]
    DeeperLab~\cite{yang2019deeperlab} & Xception-71 & & 33.8 & - & - \\
    SSAP~\cite{gao2019ssap} & ResNet-101  & \cmark & 36.5 & -    & - \\
    PCV~\cite{Wang_2020_CVPR} & ResNet-50 &        & 37.5 & 40.0 & 33.7 \\
    Panoptic-DeepLab~\cite{cheng2019panoptic} & Xception-71 & \cmark & 41.2 & 44.9 & 35.7 \\
    Axial-DeepLab~\cite{wang2020axial} & Axial-ResNet-L  & \cmark & 43.9 & 48.6 & 36.8 \\
    \midrule\midrule
    Panoptic-DeepLab & SWideRNet-(1, 1, 4) & & 45.3 & 51.5 & 36.1 \\
    Panoptic-DeepLab & SWideRNet-(1, 1, 4) & \cmark & 45.8 & 51.0 & 38.0 \\
    \bottomrule[0.1em]
  \end{tabular}
  }
  \caption{COCO {\it val} set. {\bf MS:} Multi-scale inputs. }
  \label{tab:coco_val}
  \vspace{-2mm}
\end{table}

\begin{table}[!t]
    \setlength{\tabcolsep}{0.9em}
        \centering
        \scalebox{0.64}{
        \begin{tabular}{l|c|c|ccc}
            \toprule[0.2em]
            Method & Backbone & MS & PQ & PQ\textsuperscript{Th} & PQ\textsuperscript{St} \\
            \toprule[0.2em]
            \multicolumn{6}{c}{Top-down (Box-based) panoptic segmentation methods} \\
            \midrule
            TASCNet~\cite{li2018learning} & ResNet-50 & & 40.7 & 47.0 & 31.0 \\
            Panoptic-FPN~\cite{kirillov2019panoptic} & ResNet-101 & & 40.9 & 48.3 & 29.7  \\
            DETR~\cite{carion2020end} & ResNet-101 & & 46.0 & - & - \\
            AUNet~\cite{li2018attention} & ResNeXt-152 & & 46.5 & 55.8 & 32.5 \\
            UPSNet~\cite{xiong2019upsnet} & DCN-101 \cite{dai2017deformable} & \cmark & 46.6 & 53.2 & 36.7 \\
            Li~\etal~\cite{li2020unifying} & DCN-101 \cite{dai2017deformable} & & 47.2 & 53.5 & 37.7 \\
            SpatialFlow~\cite{chen2019spatialflow} & DCN-101 \cite{dai2017deformable} & \cmark & 47.3 & 53.5 & 37.9 \\
            SOGNet~\cite{yang2019sognet} & DCN-101 \cite{dai2017deformable} & \cmark & 47.8 & - & - \\
            DetectoRS~\cite{qiao2020detectors} & ResNeXt-101 & \cmark & 49.6 & 57.8 & 37.1 \\
            \midrule
            \multicolumn{6}{c}{Bottom-up (Box-free) panoptic segmentation methods} \\
            \midrule
            DeeperLab~\cite{yang2019deeperlab} & Xception-71 & & 34.3 & 37.5 & 29.6 \\
            SSAP~\cite{gao2019ssap} & ResNet-101 & \cmark & 36.9 & 40.1 & 32.0 \\
            PCV~\cite{Wang_2020_CVPR} & ResNet-50 &  & 37.7 & 40.7 & 33.1 \\
            Panoptic-DeepLab~\cite{cheng2019panoptic} & Xception-71 & \cmark & 41.4 & 45.1 & 35.9 \\
            AdaptIS~\cite{sofiiuk2019adaptis} & ResNeXt-101 & \cmark & 42.8 & 53.2 & 36.7 \\
            Axial-DeepLab-L~\cite{wang2020axial} & Axial-ResNet-L & \cmark & 44.2 & 49.2 & 36.8 \\
            \midrule \midrule
            Panoptic-DeepLab & SWideRNet-(1, 1, 4) & \cmark & 46.7 & 52.2 & 38.3 \\
            \bottomrule[0.1em]
        \end{tabular}
        }
        \caption{COCO test-dev set. {\bf MS:} Multi-scale inputs.}
        \label{tab:coco_test}
    \end{table}

    \begin{table}[!t]
        \scalebox{0.7}{
        \begin{tabular}{l|c|c|ccc}
        \toprule[0.2em]
            Model & Extra Data & MS & PQ & AP & mIoU \\
            \toprule[0.2em]
            SSAP~\cite{gao2019ssap} & & \cmark & 61.1 & 37.3 & - \\
            AdaptIS~\cite{sofiiuk2019adaptis} & & \cmark & 62.0 & 36.3 & 79.2 \\
            Panoptic-DeepLab w/ Xception-71~\cite{cheng2019panoptic} & & & 63.0 & 35.3 & 80.5 \\
            Panoptic-DeepLab w/ Xception-71~\cite{cheng2019panoptic} & & \cmark & 64.1 & 38.5 & 81.5 \\
            Axial-DeepLab-XL~\cite{wang2020axial} & & & 64.4 & 36.7 & 80.6 \\
            Axial-DeepLab-XL~\cite{wang2020axial} & & \cmark & 65.1 & 39.0 & 81.1 \\
            \midrule
            Panoptic-DeepLab w/ SWideRNet-(1, 1, 4.5) & & & 66.4 & 40.1 & 82.2 \\
            Panoptic-DeepLab w/ SWideRNet-(1, 1, 4.5) & & \cmark & 67.5 & 43.9 & 82.9 \\
            \midrule \midrule
            SpatialFlow~\cite{chen2019spatialflow} & COCO & \cmark & 62.5 & - & - \\
            Seamless~\cite{porzi2019seamless} & MV &  & 65.0 & - & 80.7 \\
            Panoptic-DeepLab w/ Xception-71~\cite{cheng2019panoptic} & MV & & 65.3 & 38.8 & 82.5 \\
            Panoptic-DeepLab w/ Xception-71~\cite{cheng2019panoptic} & MV & \cmark & 67.0 & 42.5 & 83.1 \\
            Axial-DeepLab-XL~\cite{wang2020axial} & MV & & 67.8 & 41.9 & 84.2 \\
            Axial-DeepLab-XL~\cite{wang2020axial} & MV & \cmark & 68.5 & 44.2 & 84.6\\
            \midrule
            Panoptic-DeepLab w/ SWideRNet-(1, 1, 4.5) & MV & & 68.5 & 42.8 & 84.6 \\
            Panoptic-DeepLab w/ SWideRNet-(1, 1, 4.5) & MV & \cmark & 69.6 & 46.8 & 85.3 \\
            \bottomrule[0.1em]
        \end{tabular}
        }
        \caption{Cityscapes val set. {\bf MS:} Multi-scale inputs. {\bf C:} Cityscapes coarse annotation. {\bf V:} Cityscapes video. {\bf MV:} Mapillary Vistas.}
        \label{tab:cityscapes_val}
    \end{table}

     \begin{table}[!t]
        \scalebox{0.75}{
        \begin{tabular}{l|c|ccc}
        \toprule[0.2em]
            Model & Extra Data & PQ & AP & mIoU \\
            \toprule[0.2em]
            DecoupleSegNet~\cite{li2020improving} & & - & - & 82.8 \\
            Zhu~\etal~\cite{zhu2019improving} & C, V, MV & - & - & 83.5 \\
            DecoupleSegNet~\cite{li2020improving} & MV & - & - & 83.7 \\
            HRNetV2 + OCR + SegFix~\cite{yuan2020object,yuan2020segfix,sun2019high} & C, MV & - & - & 84.5 \\
            \midrule
            AdaptIS~\cite{sofiiuk2019adaptis} & & - & 32.5 & - \\
            LevelSet R-CNN~\cite{homayounfar2020levelset} & & - & 33.3 & - \\
            UPSNet~\cite{xiong2019upsnet} & COCO & - & 33.0 & - \\
            PANet~\cite{liu2018path} & COCO & - & 36.4 & - \\
            LevelSet R-CNN~\cite{homayounfar2020levelset} & COCO & - & 40.0 & - \\
            PolyTransform~\cite{liang2019polytransform} & COCO & - & 40.1 & \\
            \midrule
            SSAP~\cite{gao2019ssap} & & 58.9 & 32.7 & - \\
            Li~\etal\cite{li2020unifying} &  & 61.0 & - & - \\
            Panoptic-DeepLab w/ Xception-71~\cite{cheng2019panoptic} & & 62.3 & 34.6 & 79.4 \\
            Axial-DeepLab-XL~\cite{wang2020axial} & & 62.8 & 34.0 & 79.9 \\
            TASCNet~\cite{li2018learning} & COCO & 60.7 & - & - \\
            Seamless~\cite{porzi2019seamless} & MV & 62.6 & - & - \\
            Li~\etal\cite{li2020unifying} & COCO & 63.3 & - & - \\
            Panoptic-DeepLab w/ Xception-71~\cite{cheng2019panoptic} & MV & 65.5 & 39.0 & 84.2 \\
            Axial-DeepLab-XL~\cite{wang2020axial} & MV & 66.6 & 39.6 & 84.1 \\
            Naive-Student~\cite{chen2020naive} & C$\dagger$, V, MV & 67.8 & 42.6 & 85.2 \\
            \midrule \midrule
            Panoptic-DeepLab w/ SWideRNet-(1, 1, 4.5) & & 64.8 & 38.0 & 80.4 \\
            Panoptic-DeepLab w/ SWideRNet-(1, 1, 4.5) & MV & 67.8 & 42.2 & 84.1 \\
            Panoptic-DeepLab w/ SWideRNet-(1, 1, 4.5) & C$\dagger$, V, MV & 68.5 & 43.4 & 85.1 \\
            \bottomrule[0.1em]
        \end{tabular}
        }
        \caption{Cityscapes test set. {\bf C:} Cityscapes coarse annotation. {\bf C$\dagger$:} Cityscapes coarse images with pseudo-labels. {\bf V:} Cityscapes video. {\bf MV:} Mapillary Vistas.}
        \label{tab:cityscapes_test}
    \end{table} 

\begin{table}[!t]
    \centering
    \scalebox{0.68}{
    \begin{tabular}{l|c|ccccc}
        \toprule[0.2em]
        Method & MS & PQ & PQ\textsuperscript{Th} & PQ\textsuperscript{St} & AP & mIoU \\
        \toprule[0.2em]
        \multicolumn{7}{c}{Top-down (Box-based) panoptic segmentation methods} \\
        \midrule
        TASCNet \cite{li2018learning} & & 32.6 & 31.1 & 34.4 & 18.5 & - \\
        TASCNet \cite{li2018learning} & \cmark & 34.3 & 34.8 & 33.6 & 20.4 & - \\
        Seamless \cite{porzi2019seamless} & & 37.7 & 33.8 & 42.9 & 16.4 & 50.4 \\
        \midrule
        \multicolumn{7}{c}{Bottom-up (Box-free) panoptic segmentation methods} \\
        \midrule
        DeeperLab \cite{yang2019deeperlab} & & 32.0 & - & - & - & 55.3 \\
        AdaptIS \cite{sofiiuk2019adaptis} & & 35.9 & 31.5 & 41.9 & - & - \\
        Panoptic-DeepLab (Auto-XL++ \cite{liu2019auto}) \cite{cheng2019panoptic} & \cmark & 40.3 & - & - & 16.9 & 57.6 \\
        Axial-DeepLab-L~\cite{wang2020axial} & & 40.1 & 32.7 & 49.8 & 16.7 & 57.6 \\
        Axial-DeepLab-L~\cite{wang2020axial} & \cmark  & 41.1 & 33.4 & 51.3 & 17.2 & 58.4 \\
        Panoptic-DeepLab (ensemble of 6 models)~\cite{cheng2019panoptic} & \cmark & 42.2 & - & - & 18.2 & 58.7 \\
        \midrule \midrule
        Panoptic-DeepLab w/ SWideRNet-(1, 1, 4.5) & & 43.7 & 38.0 & 51.2 & 21.0 & 59.4 \\
        Panoptic-DeepLab w/ SWideRNet-(1, 1, 4.5) & \cmark & 44.8 & 39.3 & 51.9 & 22.2 & 60.0 \\
        \bottomrule[0.1em]
    \end{tabular}
    }
    \caption{Mapillary Vistas validation set. {\bf MS:} Multiscale inputs.}
\label{tab:mv_val}
\end{table}

\begin{table}[!t]
  \centering
  \scalebox{0.58}{
  \begin{tabular}{l | c | c c c | c c c}
    \toprule[0.2em]
          &     & \multicolumn{3}{c}{Val set} & \multicolumn{3}{c}{Test set} \\
    Model & MS  & PQ (\%) & mIoU & Pixel-Acc & mIoU & Pixel-Acc & Score \\
    \toprule[0.2em]
    BGRNet~\cite{wu2020bidirectional} &  & 31.8 & - & - & - & - & - \\
    Auto-Panoptic~\cite{wu2020auto}    &  & 32.4 & - & - & - & - & - \\
    PSPNet (single-model)~\cite{zhao2017pyramid} & \cmark & - & 44.94 & 81.69 & - & - & 55.38 \\
    PSPNet (ensemble-model)~\cite{zhao2017pyramid} & \cmark & - & - & - & - & - & 57.21 \\
    CPN~\cite{yu2020context} & \cmark & - & 46.27 & 81.85 & - & - & - \\
    \midrule\midrule
    {\small Panoptic-DeepLab w/ SWideRNet-(1, 1, 4)} &        & 37.53 & 49.37 & 83.50 & - & - & - \\
     {\small Panoptic-DeepLab w/ SWideRNet-(1, 1, 4)} & \cmark & 37.86 & 49.96 & 83.78 & 42.09 & 76.19 & 59.14   \\
     \midrule
     {\small Panoptic-DeepLab w/ SWideRNet-(1, 1.5, 3)} &        & 36.93 & 49.63 & 83.71 & - & - & - \\
     {\small Panoptic-DeepLab w/ SWideRNet-(1, 1.5, 3)} & \cmark & 37.41 & 50.35 & 84.02 & 40.47 & 75.21 & 57.84 \\
    \bottomrule[0.1em]
  \end{tabular}
  }
  \caption{ADE20K. {\bf MS:} Multiscale inputs. {\bf Score:} Average of mIoU and Pixel-Accuracy.}
  \label{tab:ade20k_val}
  \vspace{-2mm}
\end{table}

\begin{figure*}[!t]
\begin{center}
\bgroup 
 \def\arraystretch{0.2} 
 \setlength\tabcolsep{0.5pt}
\begin{tabular}{ccccccccc}
\includegraphics[width=0.11\linewidth]{figures/vis/coco/000070_image.png} &
\includegraphics[width=0.11\linewidth]{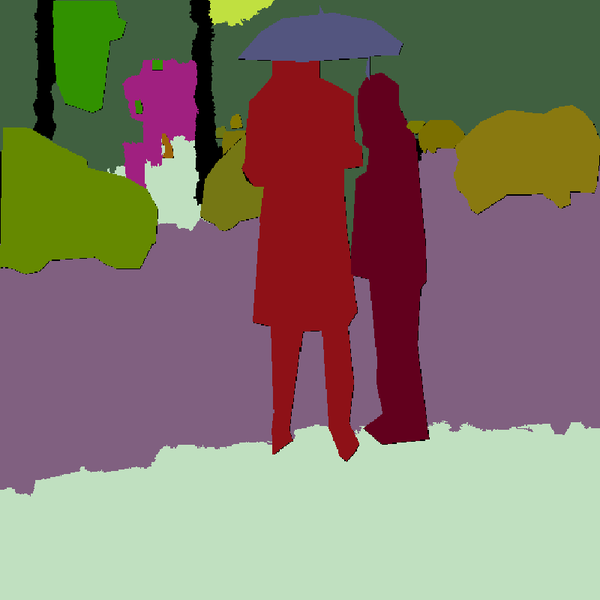} &
\includegraphics[width=0.11\linewidth]{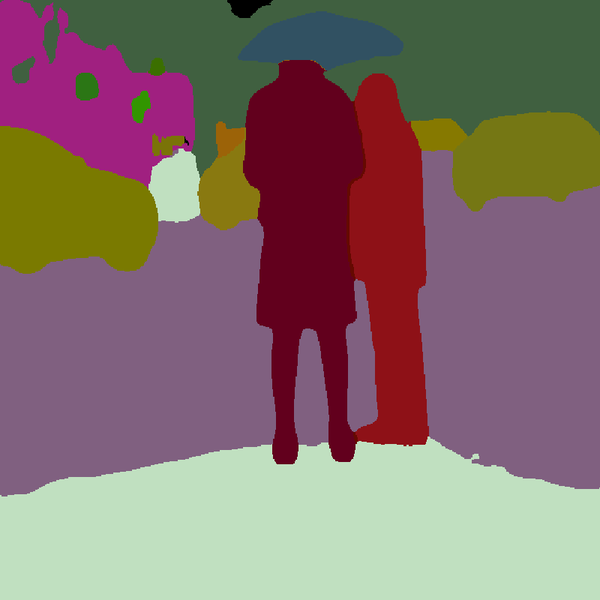} &
\includegraphics[width=0.11\linewidth]{figures/vis/coco/000080_image.png} &
\includegraphics[width=0.11\linewidth]{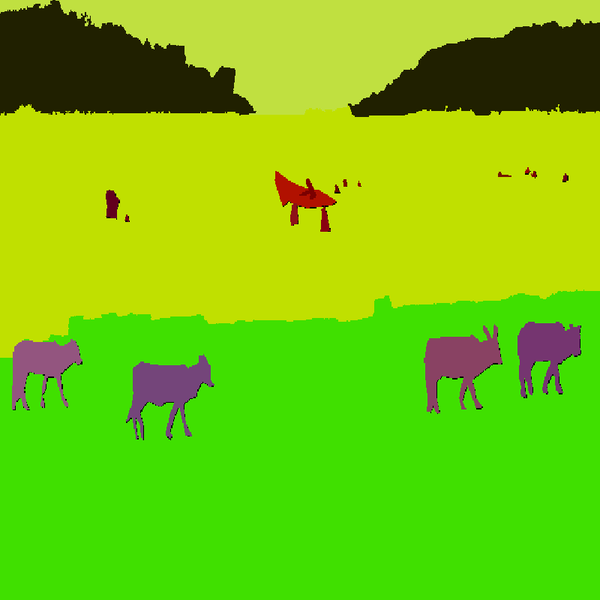} &
\includegraphics[width=0.11\linewidth]{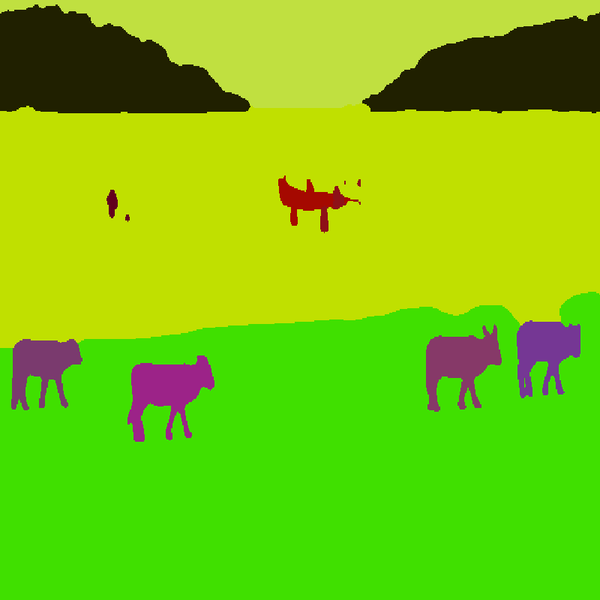} &
\includegraphics[width=0.11\linewidth]{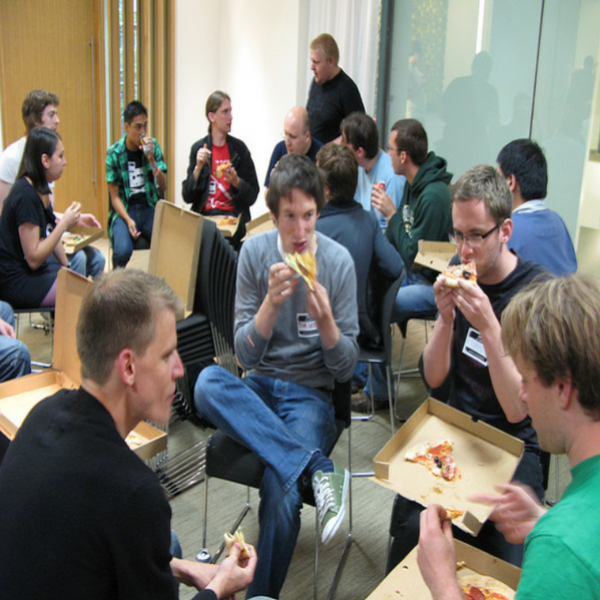} &
\includegraphics[width=0.11\linewidth]{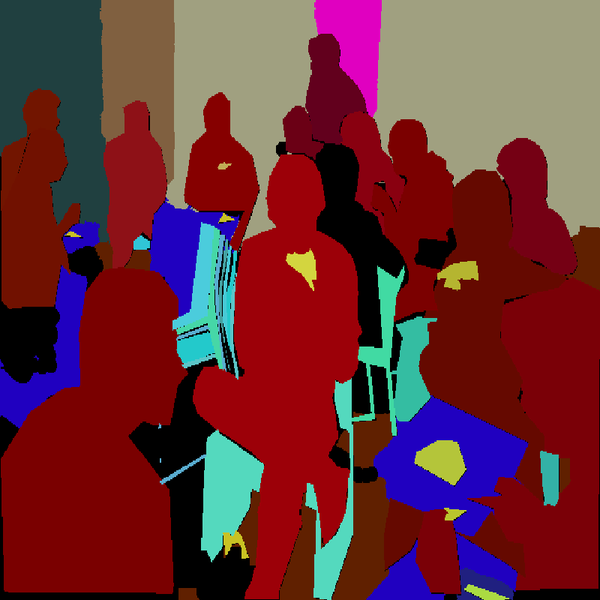} &
\includegraphics[width=0.11\linewidth]{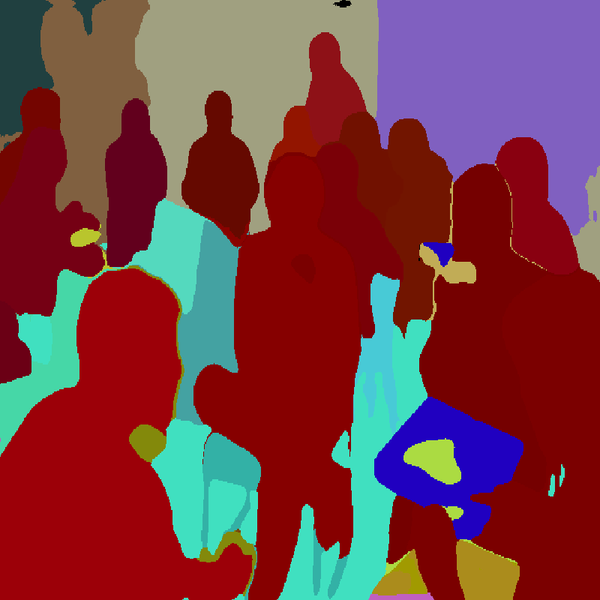} \\
\includegraphics[width=0.11\linewidth]{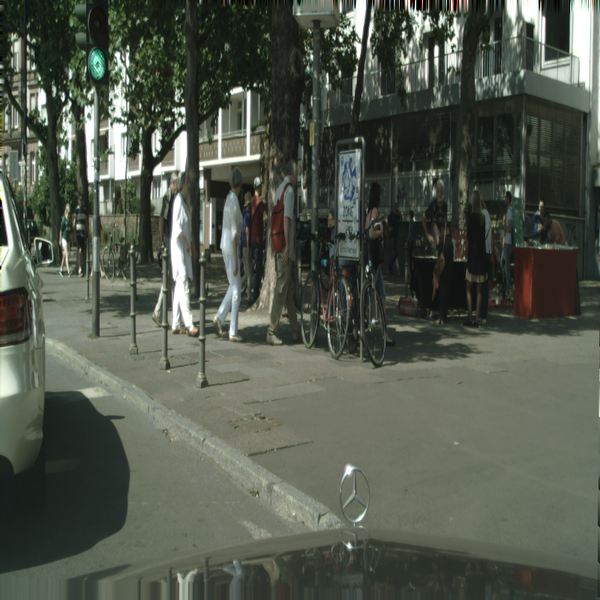} &
\includegraphics[width=0.11\linewidth]{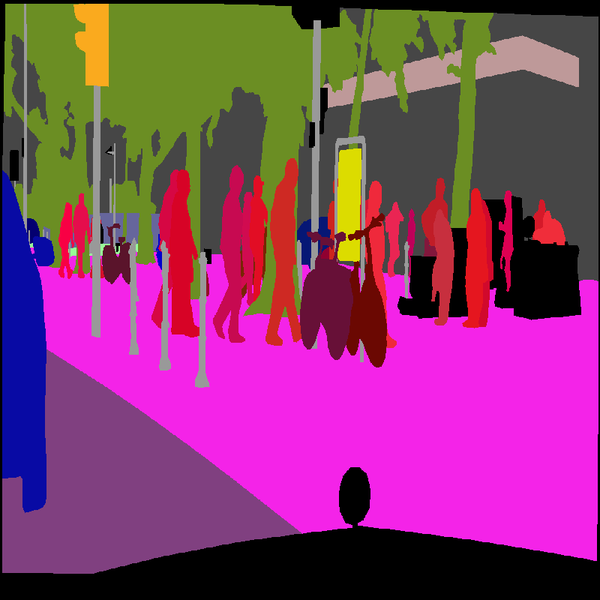} &
\includegraphics[width=0.11\linewidth]{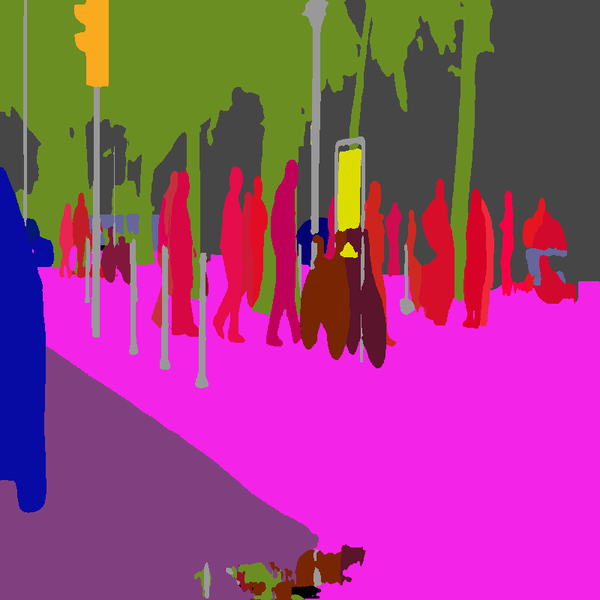} &
\includegraphics[width=0.11\linewidth]{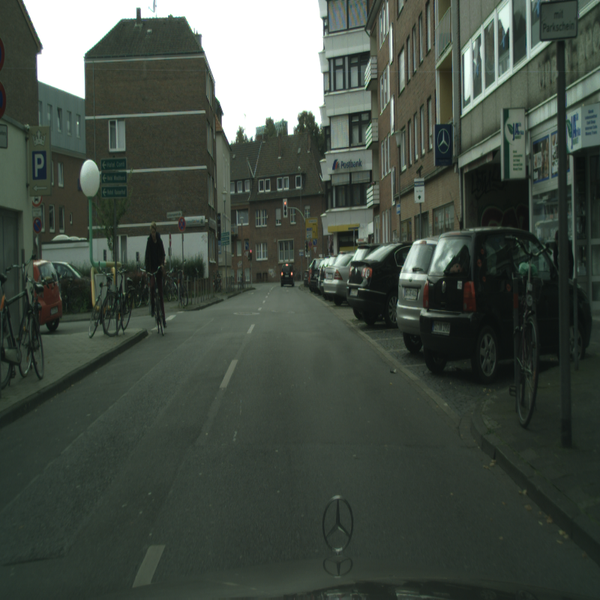} &
\includegraphics[width=0.11\linewidth]{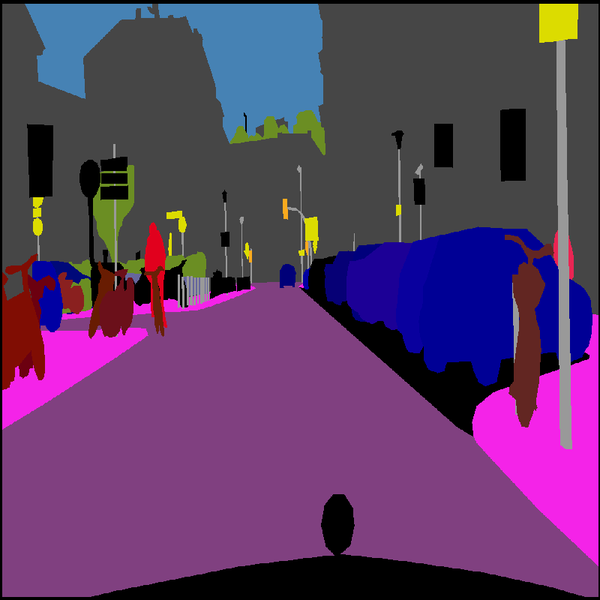} &
\includegraphics[width=0.11\linewidth]{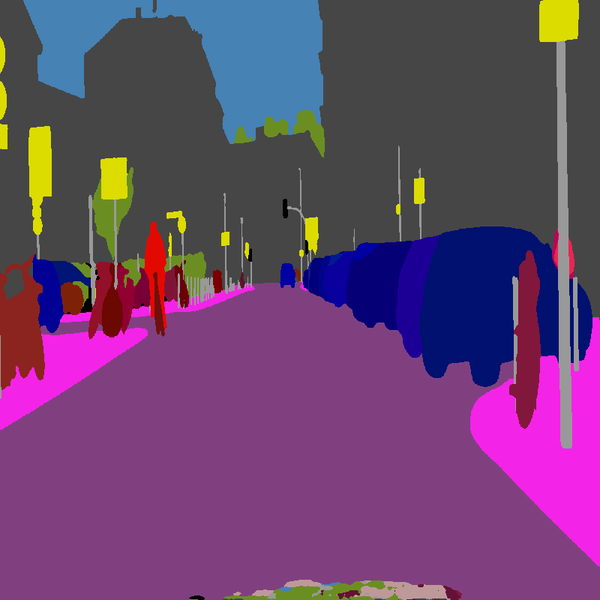} &
\includegraphics[width=0.11\linewidth]{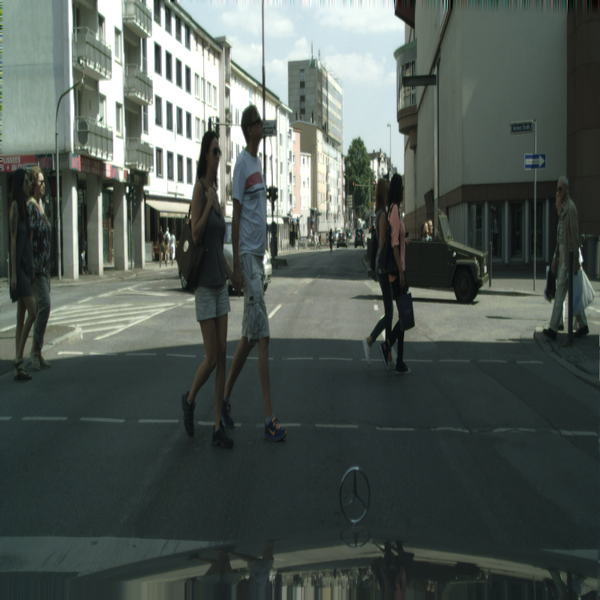}  &
\includegraphics[width=0.11\linewidth]{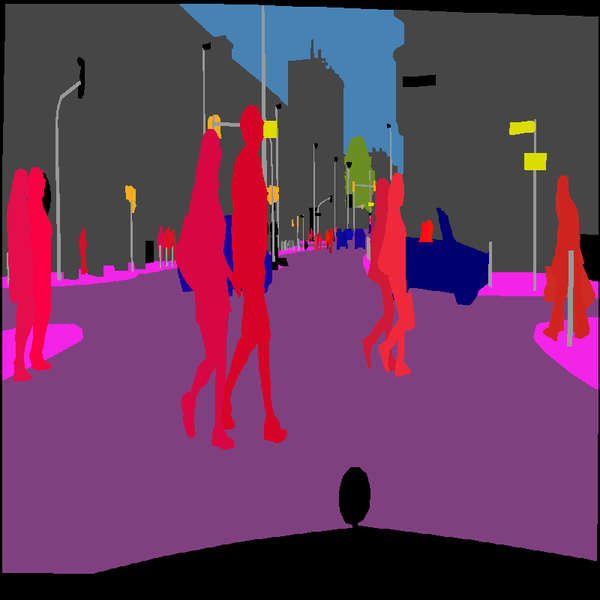} &
\includegraphics[width=0.11\linewidth]{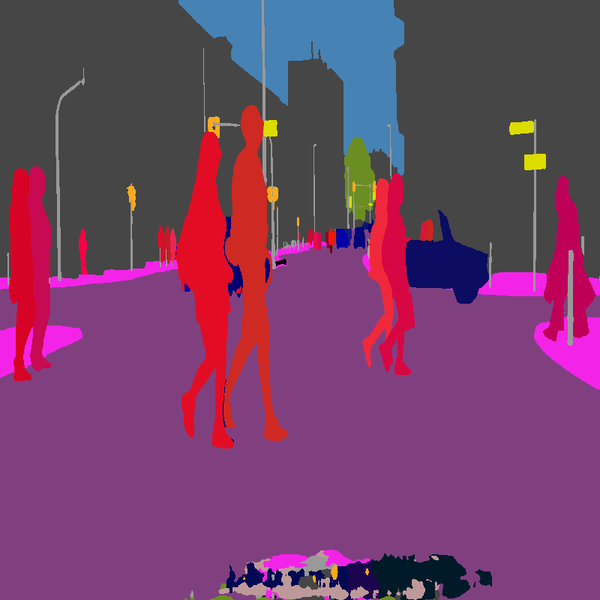} \\
\includegraphics[width=0.11\linewidth]{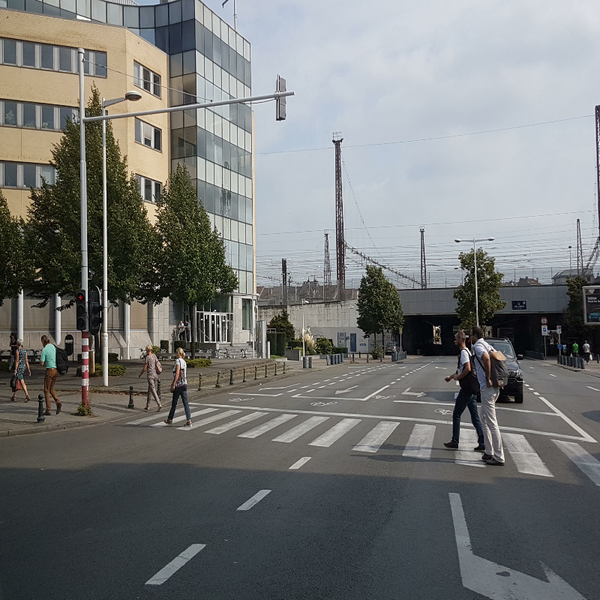} &
\includegraphics[width=0.11\linewidth]{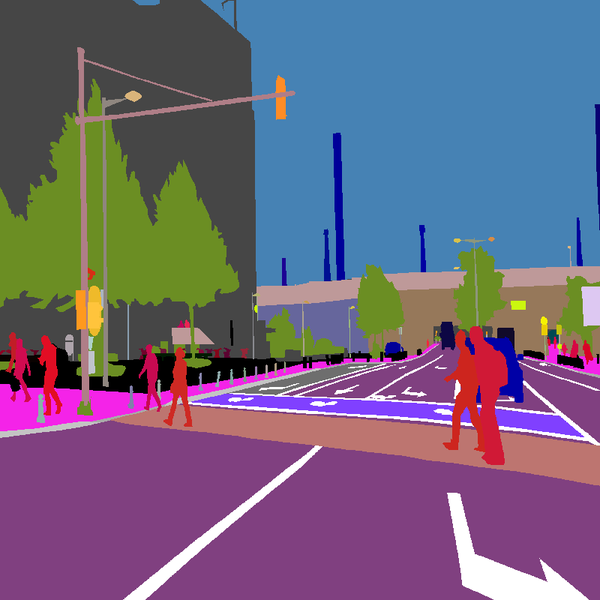} &
\includegraphics[width=0.11\linewidth]{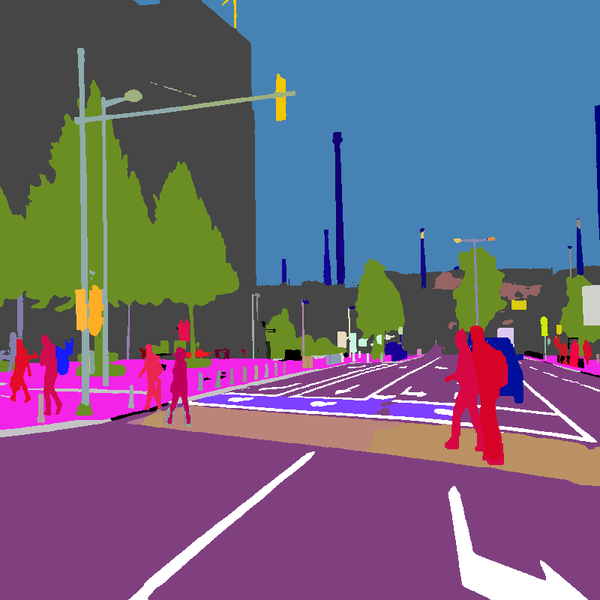} &
\includegraphics[width=0.11\linewidth]{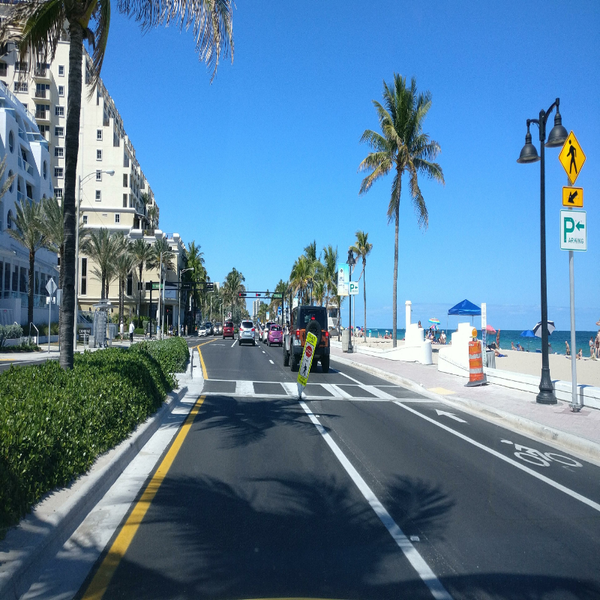} &
\includegraphics[width=0.11\linewidth]{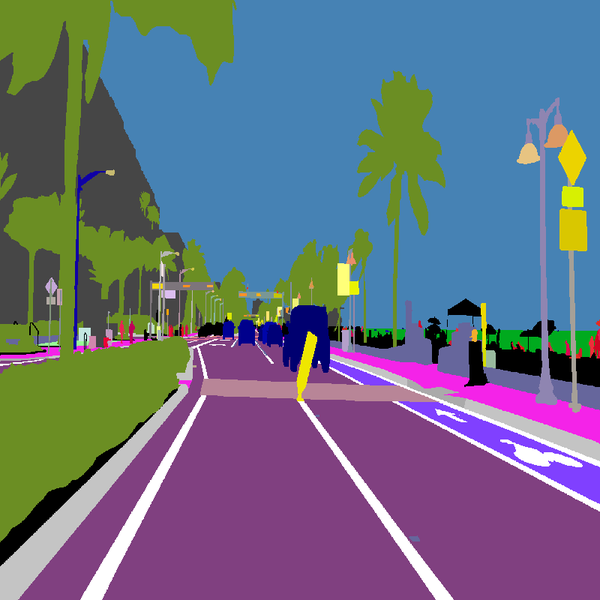} &
\includegraphics[width=0.11\linewidth]{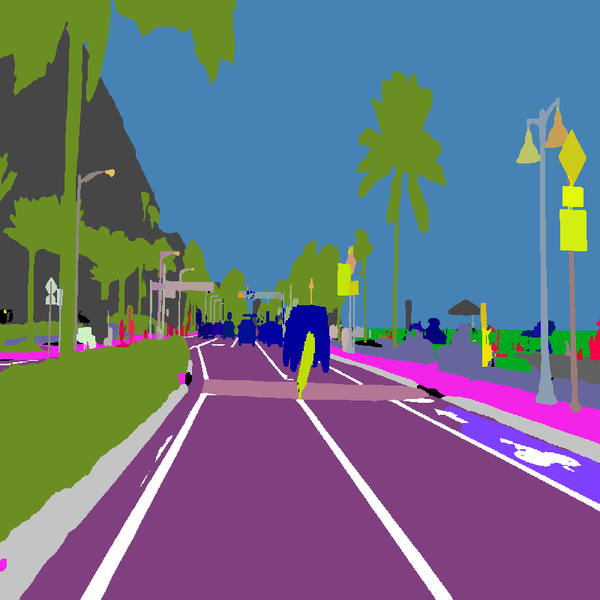} &
\includegraphics[width=0.11\linewidth]{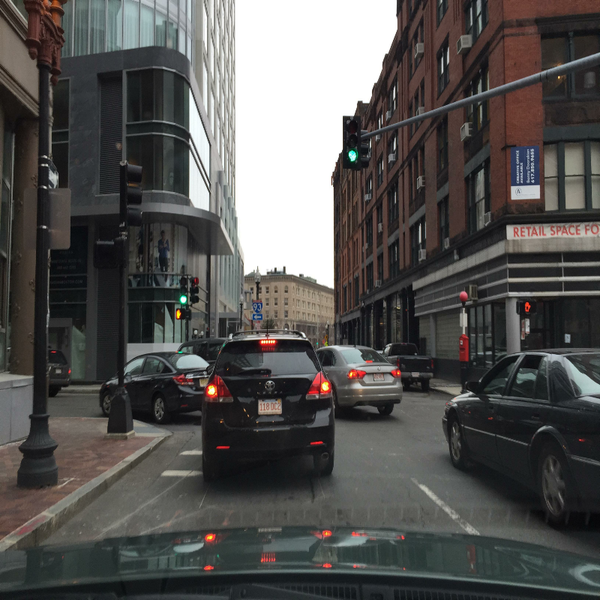} &
\includegraphics[width=0.11\linewidth]{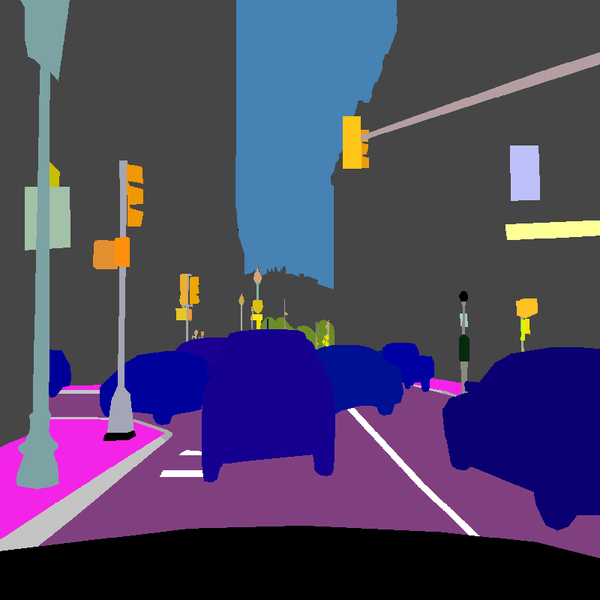} &
\includegraphics[width=0.11\linewidth]{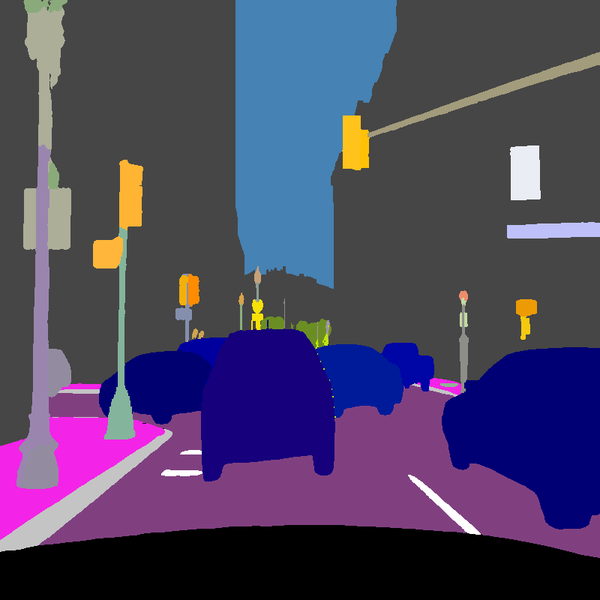} \\
\includegraphics[width=0.11\linewidth]{figures/vis/ade20k/000022_image.png} &
\includegraphics[width=0.11\linewidth]{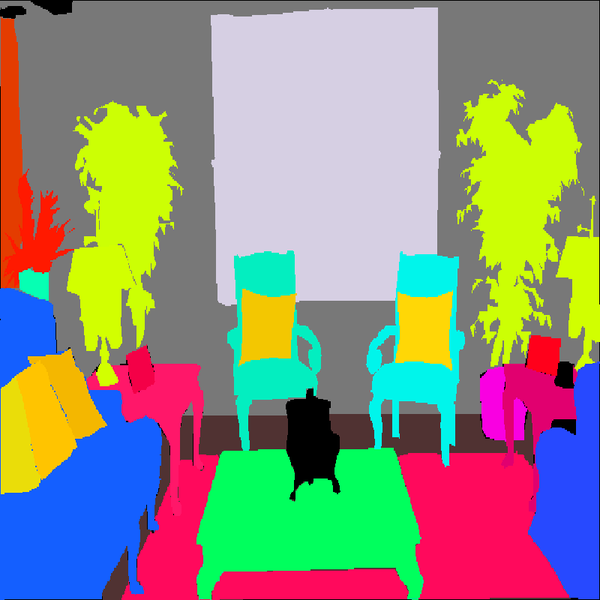} &
\includegraphics[width=0.11\linewidth]{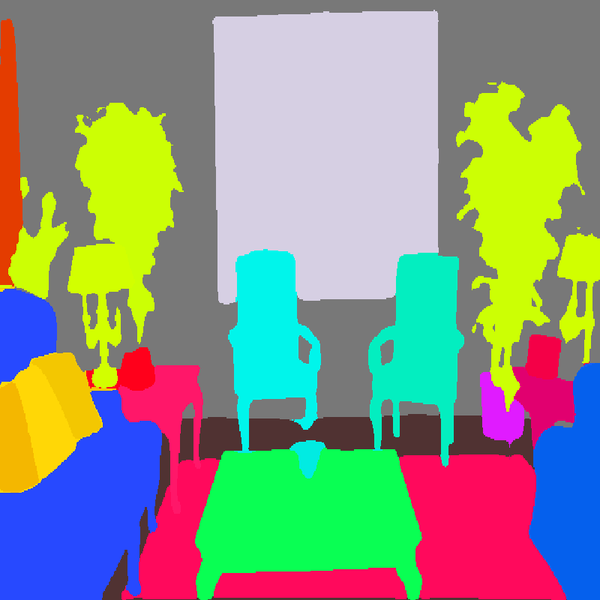} &
\includegraphics[width=0.11\linewidth]{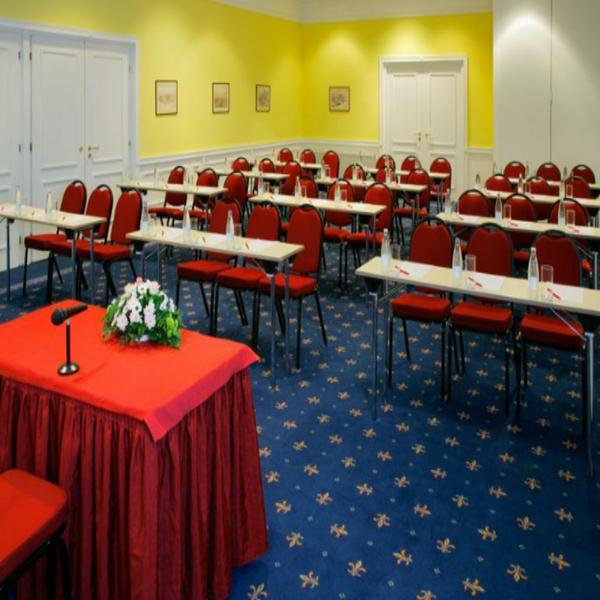} &
\includegraphics[width=0.11\linewidth]{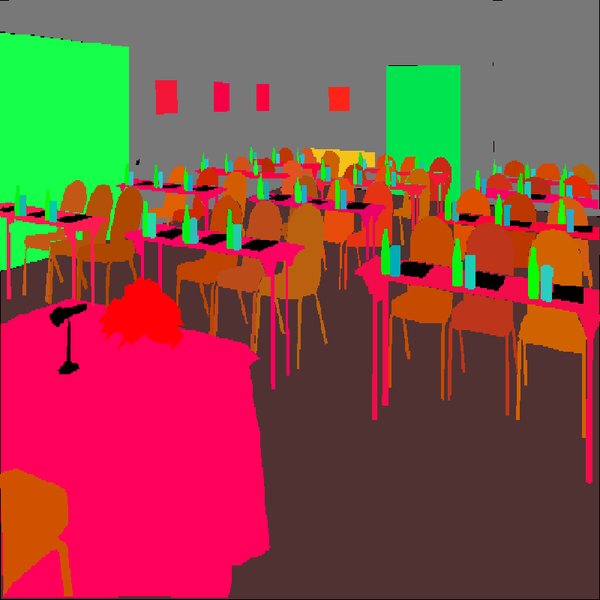} &
\includegraphics[width=0.11\linewidth]{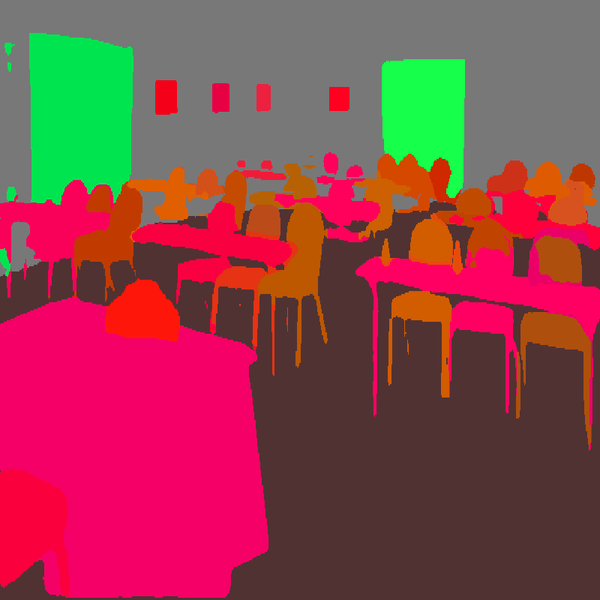} &
\includegraphics[width=0.11\linewidth]{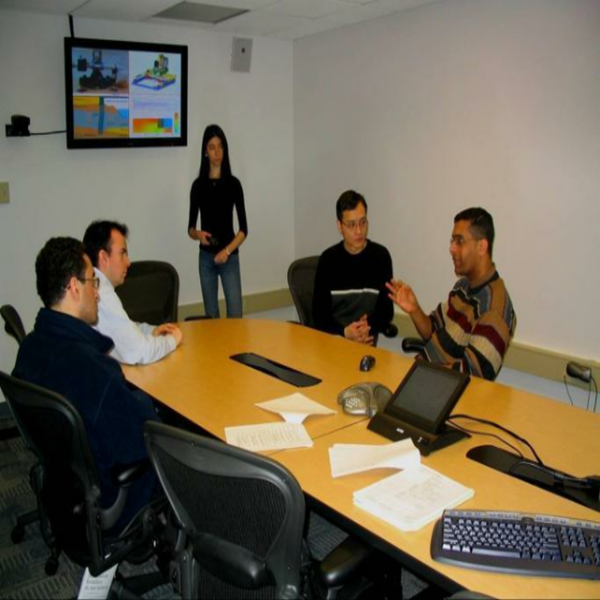} &
\includegraphics[width=0.11\linewidth]{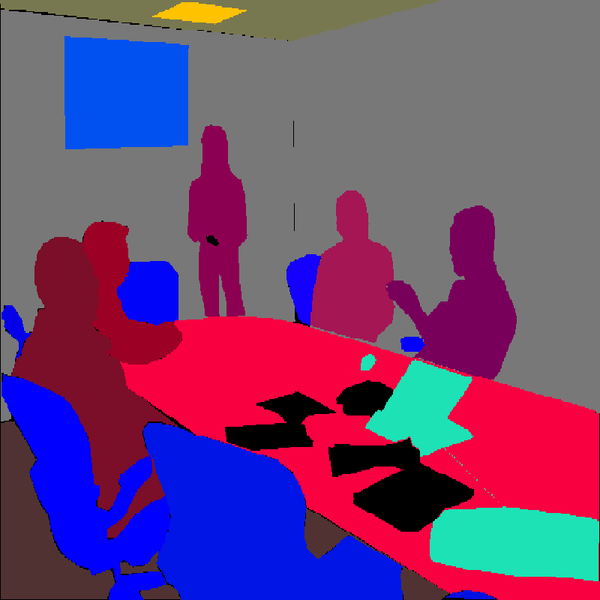} &
\includegraphics[width=0.11\linewidth]{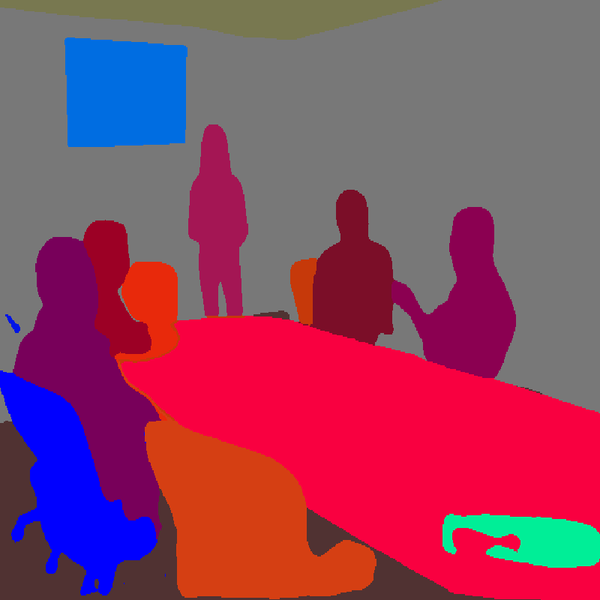} \\
\end{tabular} \egroup 
\end{center}
\caption{Our visualization results on COCO (1st row), Cityscsapes (2nd row), Mapillary Vistas (3rd row), and ADE20K (4th row). For every triple of images, we show (image, ground-truth, prediction). Our models struggle for small, thin, or heavily occluded objects.
}
\label{fig:failure}
\vspace{-2mm}
\end{figure*}

\section{Discussion}
\label{sec:discussion}

{\bf M-Adds \vs real inference speed:} We empirically discover that M-Adds (Multiply-Adds) or FLOPs are a very rough proxy of real-world inference speed (\eg, \tabref{tab:runtime}), echoing the findings from~\cite{yang2018netadapt,tan2019mnasnet}. Therefore, it is more accurate to directly measure the inference latency on the target device~\cite{dong2018dpp,cai2020once} for  comparing the speed-accuracy trade-off between different network architectures.

{\bf Model parameters:} Our SWideRNets, derived from the  Wide-ResNets~\cite{wrn2016wide,wu2019wider}, share the same issue about large model parameters. This could be potentially alleviated by pruning the networks~\cite{han2016deep,yang2018netadapt}. %

\section{Conclusion}
\label{sec:conclusion}
In this work, we present SWideRNet-$(w_1, w_2, \ell)$, a family of neural networks by scaling the width (\ie, channel size) and depth (\ie, number of layers) of Wide Residual Networks. Two search spaces are explored, where the first one contains fast model variants, and the other one contains strong architectures. Discretizing the search space allows us to employ the simple but effective grid search. As a result, we empirically identify several fast SWideRNets that attain outstanding performance in terms of speed-accuracy trade-off, and several strong SWideRNets that further advance sate-of-the-art results on several public benchmarks. 

{\bf Acknowledgments }
We would like to thank the discussions with Yukun Zhu and Maxwell Collins, and the support from Google Mobile Vision team.

{\small
\bibliographystyle{ieee_fullname}
\bibliography{refs}
}

\end{document}